\documentclass[journal]{IEEEtran}
\usepackage[utf8]{inputenc}
\usepackage{color}
\usepackage{xcolor}
\usepackage{array}
\usepackage{verbatim}
\usepackage{amsmath}
\usepackage{amsthm}
\usepackage{amssymb}
\usepackage{graphicx}
\usepackage{longtable}
\usepackage{multirow}
\usepackage{booktabs}
\usepackage{tikz}
\usepackage{pgfplots}
\usepackage{amsthm}
\pgfplotsset{compat=1.16}

\usepackage[unicode=true,
bookmarks=false,
breaklinks=false,pdfborder={0 0 1},colorlinks=false]
{hyperref}
\hypersetup{
	colorlinks,bookmarksopen,bookmarksnumbered,citecolor=blue,urlcolor=blue}
\usepackage{cite}

\usepackage{lipsum}
\usepackage{mathtools}
\usepackage{float}
\usepackage{cuted}
\providecommand{\tabularnewline}{\\}
\usepackage{algorithmic}
\usepackage{longtable}
\usepackage{prettyref}

\floatstyle{ruled}
\newfloat{algorithm}{tbp}{loa}
\providecommand{\algorithmname}{Algorithm}
\floatname{algorithm}{\protect\algorithmname}

\makeatletter
\let\oldforeign@language\foreign@language
\DeclareRobustCommand{\foreign@language}[1]{%
	\lowercase{\oldforeign@language{#1}}}

\let\oldforeign@language\foreign@language
\DeclareRobustCommand{\foreign@language}[1]{%
	\lowercase{\oldforeign@language{#1}}}

\ifCLASSINFOpdf
\else
\fi

\@ifundefined{showcaptionsetup}{}{%
	\PassOptionsToPackage{caption=false}{subfig}}
\usepackage{subfig}

\usepackage{balance}

\ifCLASSINFOpdf
\else
\fi

\ifCLASSINFOpdf
\else
\fi

\def\ps@IEEEtitlepagestyle{%
	\def\@oddhead{\parbox[t][\height][t]{\textwidth}{\centering \scriptsize
			Personal use of this material is permitted. Permission from the author(s) and/or copyright holder(s), must be obtained for all other uses. Please contact us and provide details if you believe this document breaches copyrights.\\
			\noindent\makebox[\linewidth]{}
		}\hfil\hbox{}}%
	\def\@evenhead{\scriptsize\thepage \hfil \leftmark\mbox{}}%
	\def\@oddfoot{\parbox[t][\height][l]{\textwidth}{
			\vspace{-20pt}{\rule{\textwidth}{0.4pt}}\\ \footnotesize{\bf{\footnotesize\textcolor{red}{S. Dunthorne and H. A. Hashim, "TileNet: Tile-based CNN-SVM architecture for autonomous unmanned aerial systems inspection of flat roofs," Journal of Safety Science and Resilience, pp. 100364, 2026.}}} doi: \href{https://doi.org/10.1016/j.jnlssr.2026.100364}{10.1016/j.jnlssr.2026.100364}\\\\
			\noindent\makebox[\linewidth]
		}\hfil\hbox{}}%
	\def\@evenfoot{\MYfooter}}

\makeatother
\begin{document}
	\bstctlcite{IEEEexample:BSTcontrol}

\title{TileNet: Tile-Based CNN-SVM Architecture for Autonomous Unmanned Aerial Systems Inspection of Flat Roofs}

\author{Samuel Dunthorne and Hashim A. Hashim\\
	Department of Mechanical and Aerospace Engineering, Carleton University% <-this % stops a space
	\thanks{This work was supported in part by the Mitacs Accelerate Program and Roofmaster Ottawa Inc.}
	\thanks{S. Dunthorne and H. A. Hashim are with the Department of Mechanical and Aerospace Engineering, Carleton University, Ottawa, ON, K1S-5B6, Canada (e-mail: Hhashim@carleton.ca).}
}

% \markboth{IEEE TRANSACTIONS ON INTELLIGENT TRANSPORTATION SYSTEMS, \today}{Hashim \MakeLowercase{\textit{et al.}}: Landmark and IMU Data Fusion: Systematic Convergence Geometric Nonlinear Observer for SLAM and Velocity Bias}

\markboth{}{S. Dunthorne \MakeLowercase{\textit{et al.}}: A SegNet-Driven DL Architecture for Autonomous UASs Inspection of Flat Roofs}

\maketitle
\begin{abstract}
	Flat roofs are among the most influential components of the building
	envelope, governing both structural performance and thermal efficiency,
	and thereby contributing directly to household energy consumption,
	carbon emissions, and long-term environmental sustainability. Timely
	detection of roof defects is essential for reducing heating and cooling
	losses, preventing moisture-driven degradation such as mold growth,
	and supporting national climate-change mitigation goals. This paper
	presents a real-time, Unmanned Aerial System (UAS)-based deep learning
	framework that autonomously detects defects using live imagery captured
	during dual-altitude aerial passes. The multi-resolution flight strategy
	is designed to aid the identification of both small,
		fine-scale defects and larger structural issues, enabling more comprehensive
	assessments. To meet the strict computational and power constraints
	of embedded UAS hardware, the proposed framework integrates a tile-based
	architecture with a lightweight Convolution Neural Network-Support
	Vector Machine (CNN-SVM) classifier \textcolor{black}{designed} for low-latency onboard
	inference. \textcolor{black}{Each image tile is classified as defective or non-defective, and the model performs tile-level binary classification rather than pixel-wise semantic segmentation}. The final 
	model-comprising five convolutional layers and four dense layers\textcolor{black}{, the last a linear SVM head,} achieved a mean test accuracy of
		$94.4\%$ ($95\%$ confidence interval $\pm0.4\%$ over three seeds)
		on a \textcolor{black}{photo-level split} ($43,383$ training, $3,869$
		validation, and $2,540$ test tiled and augmented images), outperforming
		GoogLeNet ($89.2\%$) and AlexNet ($79.8\%$). Experimental evaluations
	using real UAS imagery collected by onsite visits with DJI Matrice
	350 RTK drone demonstrate that the system supports rapid, repeatable,
	and safe roof inspections while reducing human risk, lowering operational
	costs, and enabling more sustainable building maintenance. 
\end{abstract}



% \fi

\section{Introduction}

\begin{figure*}[!t]
	\centering{}\includegraphics[width=7in]{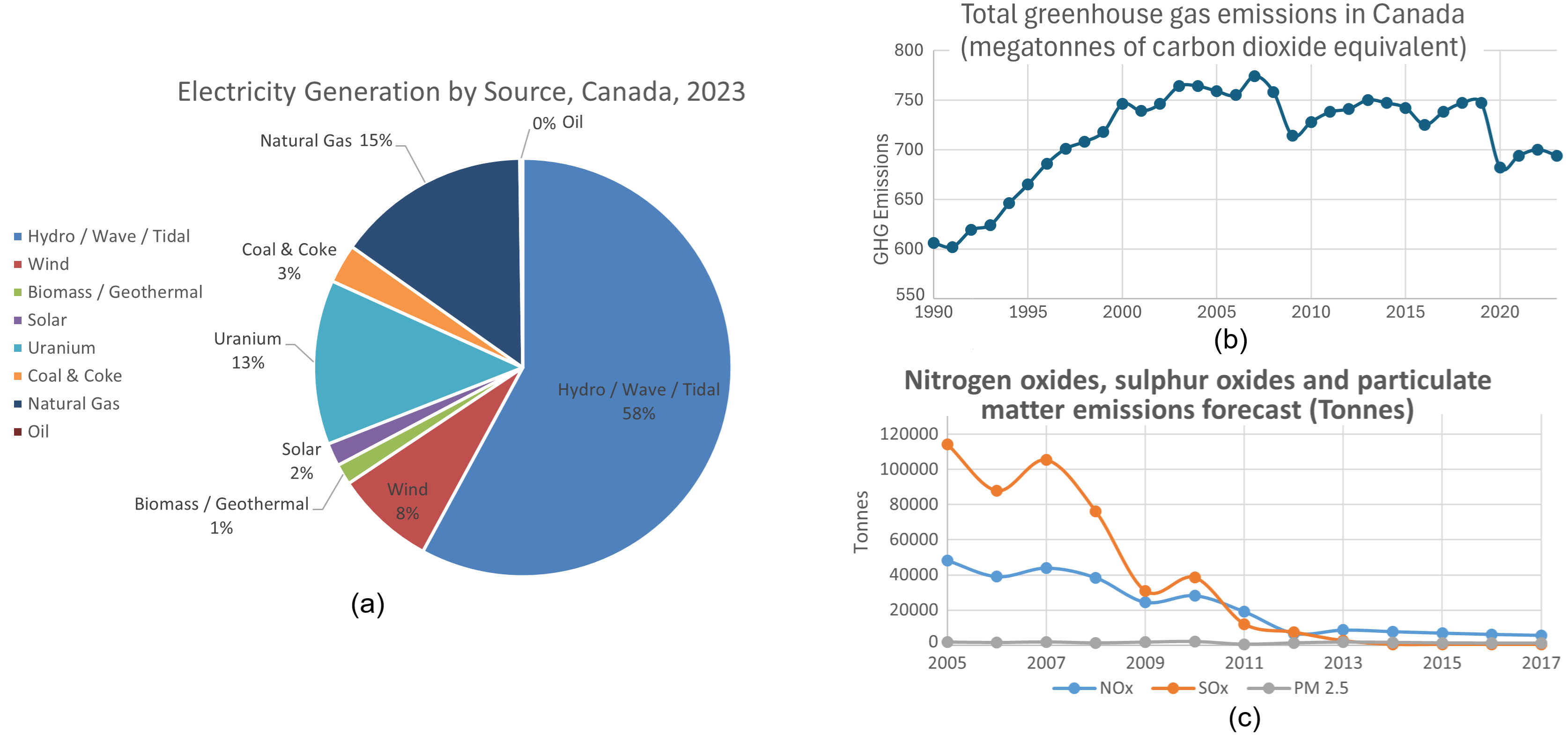}\caption{\label{fig:fig1_intro}Energy generation and its effects: (a) Electricity
		Production in Canada, $2023$ by source \cite{CanadaEnergyRegulator2023},
		(b) GreenHouse Gas (GHG) emissions in Canada, $1990$ to $2023$ \cite{EnvironmentandClimateChangeCanada2025},
		(c) Air quality trends in Ontario since $2005$ \cite{GovernmentOfOntario2025}.}
\end{figure*}

Roofing is a critical component of the building envelope, protecting
interior spaces from weather while regulating indoor climate and energy
flow. As the primary thermal and structural barrier of a building,
the roof not only prevents water intrusion but also plays a significant
role in national energy efficiency, public health, and long-term environmental
sustainability. Because roof maintenance and replacement are cyclical,
a healthy, well-performing roof directly reduces heating and cooling
demands, thereby lowering both household energy consumption and associated
carbon emissions. According to the 2025 report of Natural Resources
Canada, heating and cooling account for roughly $65\%$ of residential
energy use \cite{NaturalResourcesCanada2025}, making the thermal
integrity of roof systems central to climate-change mitigation goals
of every country. Aging or damaged roofs reduce energy efficiency
and often allow moisture ingress, which degrades insulation performance
and drives up energy demand \textcolor{black}{\cite{Tariku2023,wang2025study}}.
Solar absorption further affects thermal loads, for instance, buildings
in the province of Alberta receive on average $3.77$ $\text{kWh/m}^{2}$
and $2,506$ hours of solar energy annually \cite{MansouriKouhestani2019}.
Conventional asphalt roofs absorb substantial heat which is an effect
that worsens with age, while modern reflective membranes and green
roofs reduce heat gain, cooling buildings and lowering fossil-fuel-based
energy consumption \cite{Grant2017}. Improved roof performance thus
contributes directly to national emissions reductions and cleaner
air through lower energy usage with lower costs for energy production
and transmission, and better air quality \cite{Brook2014}. For instance,
Fig. \ref{fig:fig1_intro} presents electricity energy generation,
total GreenHouse Gas (GHG) emissions, and air quality trend in Canada
over the last two decades, revealing large amount of GHG emissions.
Given that roofs constitute one of the largest exposed surfaces on
a building, addressing any issues with the roof, maintaining, or upgrading
the insulation rating of the roof system offers a cost-effective and
high-impact strategy for reducing household carbon footprints worldwide
\cite{Bartels2024}, promoting national and global sustainability
through lower energy consumption and greenhouse gas emissions.

\begin{figure*}[!ht]
	\centering{}\includegraphics[width=7in]{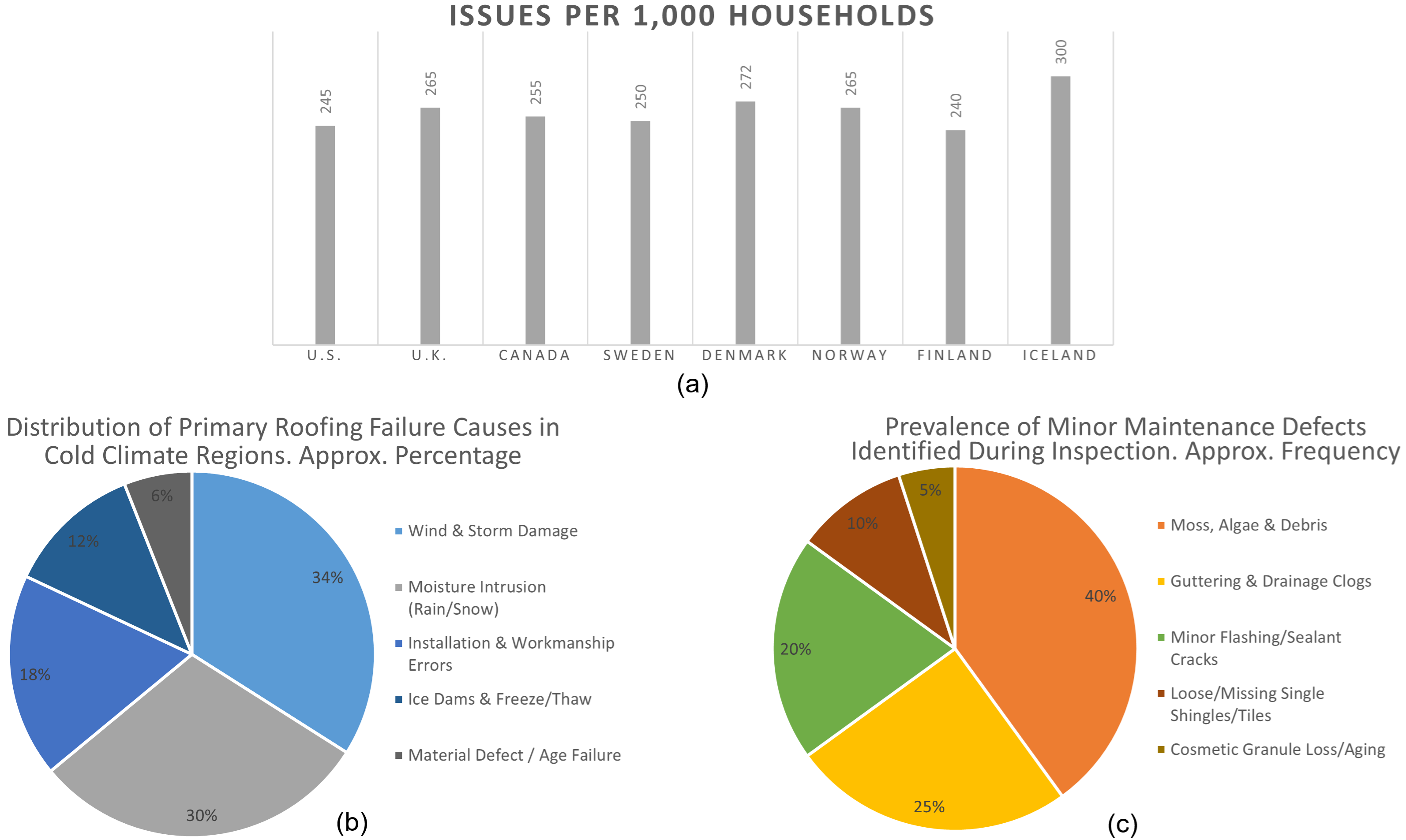}\caption{\label{fig:Fig2_roofIssues}Roofing issues broken down by: (a) occurrence
		per region in the Northern Atlantic \textcolor{black}{\cite{MinistryofHousing2024,StatisticsCanada2025,StatisticsFinland2025,StatisticsIceland2023,StatisticsNorway2025,StatisticsDenmark2025,StatisticsSweden2025,USCensusBureau2024}},
		(b) Major Issue Types \textcolor{black}{\cite{Gullbrekken2016,MinistryofHousing2024,InsuranceInformationInstitute2025}}
		and, (c) Minor Issues Types \textcolor{black}{\cite{Gullbrekken2016,MinistryofHousing2024,InsuranceInformationInstitute2025}}.}
\end{figure*}

\vspace{0.1cm}
\paragraph*{Roof Systems}

Urban areas are increasingly affected by the Urban Heat Island (UHI)
effect, where city temperatures can be $1$-$4^{\circ}$C higher than
those of surrounding rural areas under the same climate conditions
\textcolor{black}{\cite{Wang2016,Gaur2018}}. This phenomenon results from the
prevalence of artificial dark surfaces, such as asphalt, concrete,
and dark roofing materials, which exhibit low reflectivity and absorb
large amounts of solar radiation. In contrast, rural areas feature
lighter-colored surfaces, vegetation, and soil, which reflect more
sunlight and absorb less heat, leading to cooler ambient temperatures.
The UHI effect exacerbates urban energy consumption, reduces air quality,
and negatively impacts public health and sustainability, and in turn
more GHG emissions. To mitigate these impacts, urban planners increasingly
emphasize the adoption of cool roof strategies, both in new constructions
and through maintenance of existing roofs \cite{Li2014}. Cool roofs
may include vegetated (green roofs), which incorporate soil and foliage,
or reflective materials such as white Thermoplastic Polyolefin (TPO)
membranes that lower solar absorptance. These approaches reduce building
cooling loads, with studies reporting reductions of up to $34\%$
\cite{Pisello2013}. Roof systems also play a critical role in moisture
management. Uncontrolled moisture infiltration through cracks or defects
in the roofing layers can lead to mold growth, posing health risks
to occupants, particularly those with respiratory vulnerabilities
\cite{HealthCanada2023}. Wet insulation loses thermal efficiency
and becomes a breeding ground for mold and spores, which can disperse
throughout the building via HVAC systems, attic ventilation, or other
gaps between living spaces and the roof. Effective moisture control
is therefore essential not only for energy efficiency but also for
maintaining indoor air quality and overall building livability. Roofs
serve as the primary barrier of the building envelope, and their integrity
is particularly critical in regions subject to freeze-thaw cycles,
heavy precipitation, and other extreme climatic conditions \textcolor{black}{\cite{Grant2017,Wang2016,Gaur2018,Bartels2024}}.
Roofing degradation varies by geography, influenced by factors such
as high-velocity winds in North America or chronic moisture exposure
in Northern Europe. Roofing defects can generally be categorized into
two types: \textit{Major defects} involve structural or system failures,
including catastrophic wind uplift, significant moisture intrusion,
or ice dam formation, requiring immediate intervention and often incurring
substantial costs and insurance claims; and \textit{Minor defects},
typically addressed through routine maintenance rather than replacement,
may appear cosmetic, such as moss or algae growth, granular loss,
flashing deterioration, or drainage blockages, but can act as early
indicators of potential envelope failure. Data from Northern Atlantic
countries indicate that while the total number of roofing issues per
$1,000$ homes remains relatively consistent across regions, the distribution
of defect types varies depending on local climate and roof system
characteristics (Fig. \ref{fig:Fig2_roofIssues}.(b) and \ref{fig:Fig2_roofIssues}.(c)).
This highlights the universal nature of roofing challenges and the
importance of proactive maintenance strategies.

\subsection{Related Work}

Roofing system performance can be optimized through regular preventative
maintenance and timely replacement. Recent advances have extended
the lifespan of roofing systems from $10$-$20$ years to over $25$
years \cite{Aggarwal2024}. Given the ongoing challenges of human-induced
climate change, including environmental degradation, mass species
extinction, and forced migration, reductions in carbon emissions,
even at the household level, contribute to broader mitigation efforts.
In Canada, while the energy production sector has transitioned toward
cleaner sources, $19\%$ of annual energy production remains fossil-fuel-based
(see Fig. \ref{fig:fig1_intro}.(a)), and air quality improvements
in Ontario are largely attributed to the closure of coal-fired plants
and adoption of green energy sources (see Fig. \ref{fig:fig1_intro}.(b)
and (c)) \cite{GovernmentOfOntario2025}. Currently, roof inspections
are performed via visual observation, often supplemented by thermal
imaging, or through physical and destructive testing \cite{Conceicao2017}.
Physical testing involves probing membranes or flashing, while destructive
methods determine installation layers and system composition. Inspectors
rely heavily on experience and manufacturer documentation to identify
defects, and the complex variability of flat roofing systems can require
years of training. High wages and hazardous working conditions roofing accounts for roughly one-third of construction-related deaths
in the United States between $1992$ and $2009$ \cite{Dong2013} motivate the adoption of Unmanned Aerial Systems (UASs) for inspections\cite{hash2023_IJC_UAV_NavCont}.
UAS-based inspections reduce risk, save man-hours, and allow rapid,
repeatable data collection. When combined with deep learning models,
UAS can generate accurate, timely assessments of roof conditions without
relying on extensively trained human inspectors \textcolor{black}{\cite{hash2023_IJC_UAV_NavCont,hash2023_ISA_UAV_NavCont}}.
Modern imaging systems are increasingly capable of complementing or
replacing human vision through 3D machine vision and multisensor electro-optical
and infrared systems \textcolor{black}{\cite{Sakerka2012,seo2025quantitative}}.
UAS provide a versatile platform for deploying these technologies,
enabling RGB and infrared imagery capture with six degrees of freedom,
from vantage points otherwise inaccessible \textcolor{black}{\cite{Aditya2024_SegNet,hash2023_IJC_UAV_NavCont,yiugit2024automatic}}.
Deep learning models, trained on UAS-acquired data, can classify roof
surfaces and detect defects despite variations in size, shape, or
color \cite{Zahradnik2024,Santos2023}. The typical workflow involves
UAS-based image capture, local processing, and transmission of predictions
for operator review and later retrieval, a methodology proven effective
in other domains \textcolor{black}{\cite{Aditya2024_SegNet,Aditya_2024_ICDS_comprehensive}}.
\vspace{0.2cm}
\paragraph*{Need for Autonomy Detection}

Real-time autonomous roof inspection demands high model accuracy and
low-latency processing. Although modern GPUs have enhanced onboard
computation, most UAS platforms remain constrained by power, weight,
and limited capacity to process high-resolution imagery (e.g., $4,032\times3,024$
pixels). Model training is feasible on ground-based resources, but
real-time inference onboard remains challenging. Progress is further
limited by a lack of defect-labeled datasets and by environmental
factors, such as shadows, cloud cover, and precipitation that reduce
detection reliability. High-altitude imaging from satellites or manned
aircraft also lacks sufficient spatial resolution. 
	While inspections can be performed at close range, UAS-based inspections
	are typically conducted at altitudes of 5 to 25 m \cite{Fan2025}.
Deep learning performance is tied to network scale and design: larger
architectures improve feature extraction but increase processing latency,
which is critical for UAS operations. Dataset expansion and augmentation
can enable smaller, more efficient models to achieve comparable accuracy.
However, creating high-quality training data is labor-intensive and
requires domain expertise, as flat-roof defects and healthy installations
often appear visually similar.

\subsection{Modern Machine Learning}

Rooftop defect detection has traditionally relied on human-led inspections.
Recent advances in thermal imaging and UAS have expanded the tools
available to inspectors, while deep learning has rapidly emerged as
a leading strategy for defect detection across sectors such as medicine,
automotive, and electrical engineering \textcolor{black}{\cite{Santos2023,Ling2023,Aditya2024_SegNet,hash2025_RIENG_Avionics}}.
Numerous studies now demonstrate that Artificial Neural Networks can
outperform or replace manual inspection processes with high accuracy
\cite{Bhatt2021}. Human-based inspection, although valuable, is limited
by inefficiency, variability in expertise, and susceptibility to environmental
conditions. In contrast, improvements in machine learning, computational
power, and sensor technologies have broadened both the complexity
and reliability of automated defect-detection tasks \cite{Santos2023,Zahradnik2024}.
While traditional image-processing approaches are effective only under
controlled conditions, modern deep learning methods operate more robustly
across variations in lighting, noise, and surface appearance \cite{Zahradnik2024}.
A range of techniques including Generative Adversarial Networks (GANs),
Support Vector Machines (SVM), Local Binary Patterns (LBP), MobileNet
Single Shot Multibox Detector, Region-Based Convolutional Neural Networks
(CNNs), Self-Organizing Maps, Convolutional Denoising Autoencoders,
and VGG architectures have all been evaluated for defect detection
\textcolor{black}{\cite{Ling2023,Bhatt2021,Santos2023,Zahradnik2024}}. Additional
approaches such as Deep Convolutional Neural Networks (DCNN), Learning
Vector Quantization (LVQ), and Multi-Layer Perceptron (MLP) have also
shown promise \textcolor{black}{\cite{Tulbure2022,Ling2023}}. Despite methodological
diversity, CNN-based approaches remain the most consistently recommended,
though their performance is often constrained by limited datasets,
particularly for rare defect types. These techniques impose design
constraints requiring an algorithm capable of running directly on
a UAS, thereby excluding many otherwise effective machine learning
methods whose power and processing demands exceed onboard capabilities
\textcolor{black}{\cite{hash2025_RIENG_Avionics,hash2026_MFWAV}}. Although compact
compute modules are emerging, mobile hardware still falls far short
of ground-based performance, and heavier systems would interfere with
the UAS's primary mission \textcolor{black}{\cite{hash2023_IJC_UAV_NavCont,Aditya2024_SegNet,Ren2015}}.
UAS-based image collection remains superior to manned aviation or
satellite imaging due to its lower cost, manoeuvrability, and ability
to capture close-range, high-density imagery from multiple angles
and altitudes \cite{hash2023_IJC_UAV_NavCont}. Most rooftop defects
require near-surface photography, which is unattainable from high-altitude
platforms, making UAS the most suitable choice for this work. A high-quality
dataset is essential. It must comprehensively represent common defect
types and be precisely labeled, as detection is complicated by factors
such as shadows, water accumulation, and visually similar surface
conditions.

\subsection{Contribution and Structure}

This paper presents a novel deep learning framework for UAS-based
flat-roof defect detection that integrates image tiling,
lightweight CNN-SVM classification, and multi-altitude UAS imaging
with rigorous experimental validation. The key contributions are listed
as follows: 
\begin{itemize}
	\item[(a)] A tile-based neural architecture (CNN-SVM)
	that preserves high-resolution defect cues while reducing computational
	load, enabling near-real-time inference on embedded UAS hardware. 
	\item[(b)] A computation-aware CNN-SVM design with optimized hyperparameters
	and reduced feature-map dimensionality, achieving over 92\% in testing
	accuracy while remaining deployable on low-power processors. 
	\item[(c)] \textcolor{black}{A multi-resolution, dual-altitude UAS imaging
		strategy, motivated by prior multi-altitude inspection studies and adopted
		as a design rationale to target small, irregular, or partially occluded
		defects. Further work is required to isolate the benefit of this is required.}
	\item[(d)] An end-to-end operational pipeline for autonomous roof inspection,
	validated with real UAS imagery (DJI Matrice 350 RTK with the camera
	mounted to DJI Zenmuse H30T) obtained from onsite visits and designed
	to reduce human risk, lower inspection costs, and support sustainable
	building maintenance. An experimental demonstration video summarizing
	the final autonomous CNN-SVM-based framework for
	drone-based rooftop defect inspection can be accessed at \href{https://youtu.be/9RsdTeuyIJM}{https://youtu.be/9RsdTeuyIJM}. 
\end{itemize}
\vspace{0.2cm}
\paragraph*{Structure}

The remainder of this paper is organized as follows. Section \ref{sec:problemformulation}
presents the problem formulation and details the dataset preparation,
image tiling strategy, data augmentation techniques,
and the image information-density analysis. Section \ref{sec:methodology}
outlines the research methodology, including the proposed tiling
approach and the model architecture. Section \ref{sec:WorkflowChalleng}
describes the overall workflow, identifies key challenges encountered
during development, and discusses corresponding mitigation strategies.
Section \ref{sec:results} reports the experimental results of the
proposed model and provides a comparative evaluation against state-of-the-art
methods. Finally, Section \ref{sec:conclusion} concludes the paper
and highlights the main contributions.

\section{Problem formulation\label{sec:problemformulation}}

Artificial Intelligence has rapidly advanced in recent years, driven
by increased computational capabilities and widespread deployment
of machine learning applications \textcolor{black}{\cite{Aditya2024_SegNet,Ren2015,Jiao2019}}.
Neural networks, inspired by the structure and function of biological
neurons, seek to approximate aspects of the brain's processing mechanisms,
an objective that remains computationally demanding despite significant
algorithmic progress. Modern deep learning models require substantial
processing power, memory, and energy, which constrains their deployment
on mobile platforms such as UAS \textcolor{black}{\cite{hash2023_IJC_UAV_NavCont,hash2025_RIENG_Avionics}}.
Unlike stationary computing systems, UAS must remain lightweight and
energy-efficient, limiting the feasibility of large, resource-intensive
neural architectures. Consequently, developing algorithms that deliver
high accuracy while remaining computationally efficient is essential
for real-time onboard inference. This challenge illustrates the need
for alternative or streamlined machine learning approaches capable
of supporting real-time rooftop defect detection within the strict
power, mass, and processing constraints of UAS platforms. 
\begin{figure*}[!t]
	\centering{}\includegraphics[scale=0.35]{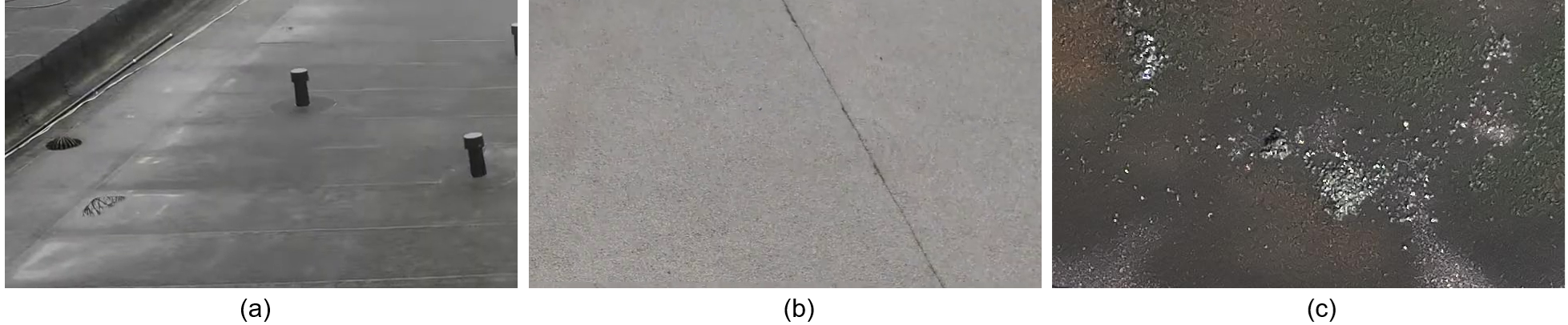}\caption{\label{fig:Fig3_enviroFactors}Environmental challenges: (a) A roof
		membrane that is damp due to recent rainfall, (b) a dry section of
		roof membrane, and (c) a section of roof membrane that is wet due
		to an issue causing prolonged standing water.}
\end{figure*}

\subsection{Dataset and Preparation\label{ssec:datasetPreparation}}

The training dataset contains some images that may work against the
goal of this project. As illustrated in Fig. \ref{fig:Fig3_enviroFactors},
certain environmental conditions, such as lighting, shadows, or recent
precipitation can cause non-defective roofs to appear visually similar
to defective ones. For example, Fig. \ref{fig:Fig3_enviroFactors}.(a)
shows a damp roof membrane that is not indicative of a defect; the
moisture likely results from recent rainfall and will dissipate under
normal conditions (e.g., warm temperatures, low humidity, solar exposure).
Flat roof systems are designed to tolerate short-term water exposure
\cite{Tariku2023}. In contrast, Fig. \ref{fig:Fig3_enviroFactors}.(c)
depicts a superficially similar but defective roof where long-term
ponding occurs due to structural degradation or improper slope. Prolonged
standing water accelerates membrane deterioration, increases the risk
of biological growth, and ultimately contributes to roof system failure
\cite{Carretero-Ayuso2016}. The visual similarity between transient,
benign moisture conditions and true defects poses a significant challenge
for supervised learning models, as both appear alike in RGB imagery.
This highlights the need for improved data differentiation strategies,
such as expanded datasets, enhanced annotation protocols, or complementary
sensing methods, to ensure robust and reliable defect classification.
Consistent with these challenges, recent work on
	flat roofs has used deep semantic segmentation to map surface stains.
	These defects have proven to be a common source of ambiguity and false
	positives in automated inspection \cite{Santos2025}.
\begin{figure*}[!t]
	\centering{}\includegraphics[scale=0.18]{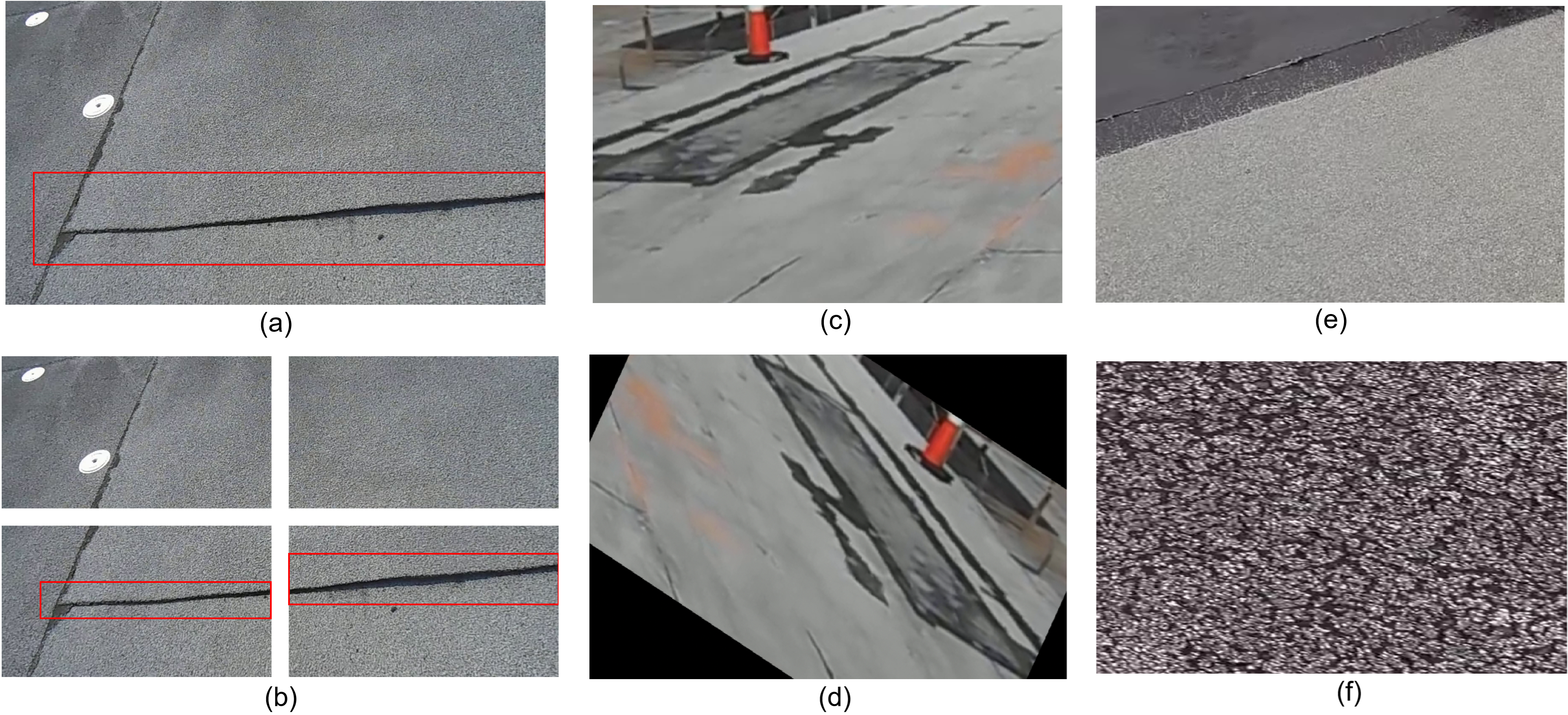}\caption{\label{fig:Fig4_SegAugZoom}Dataset preparation: (a) An image highlighting
		defects in a roof surface, (b) a visualization of the same image tiled
		into smaller subsections, (c) an unmodified image, and (d) an augmented
		version of the image, (e) a tiled image taken from
		a distance away from the roof surface, and (f) a tiled
		image close to the roof surface.}
\end{figure*}

\subsection{Image Tiling\label{ssec:image_tiling}}

\begin{figure*}[!t]
	\centering{}\includegraphics[width=1\textwidth]{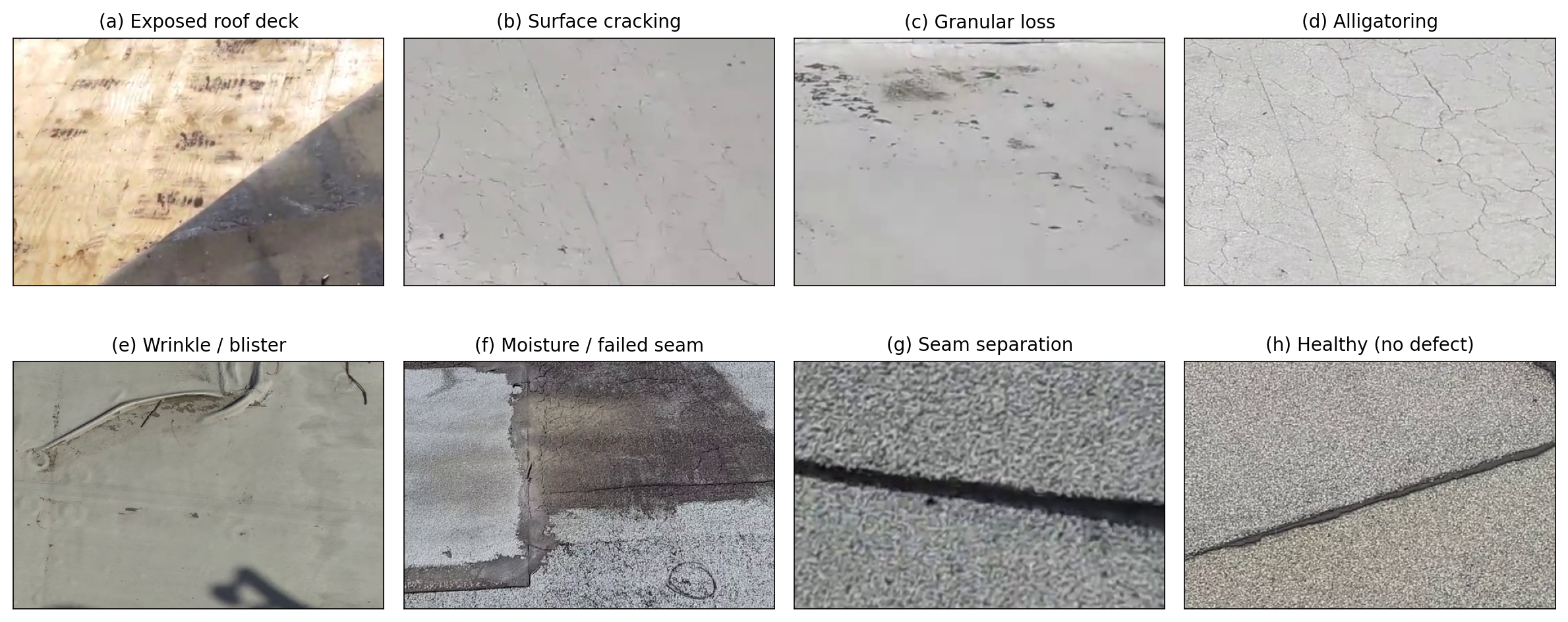}\caption{\label{fig:defectCatalog}Representative dataset
			tiles for the defect categories the CNN-SVM is trained to detect:
			(a) exposed roof deck, (b) surface cracking, (c) granular loss, (d)
			alligatoring, (e) wrinkling/blistering, (f) moisture staining and
			a failed seam, and (g) membrane seam separation; (h) shows a healthy
			membrane for contrast. All tiles are drawn from the labelled dataset,
			and the binary defect/no-defect label aggregates these modified-bitumen
			failure modes.}
\end{figure*}

At the outset of this research, no publicly available, labeled dataset
suitable for rooftop defect classification was identified. Unlike
many contemporary machine learning domains that benefit from extensive,
well-structured datasets, rooftop defect detection remains a relatively
unexplored problem. Existing object-detection frameworks such as YOLOv4
and R-CNN \textcolor{black}{\cite{Ren2015,Bochkovskiy2020}} can reduce some annotation
effort, but they still require substantial manual labeling and do
not circumvent the need for a domain-specific dataset. Consequently,
a custom dataset was assembled and annotated for this study. Images
were sourced from limited online repositories as well as collected
directly using a UAS (DJI Matrice 350 RTK drone via onsite visits).
Online imagery, though useful, was sparse and heterogeneous, leading
to inconsistencies in data quality and resolution ($1,280\times720$
to $1,920\times1,080$ pixels). In contrast, UAS-acquired images provided
high-resolution data ($4,032\times3,024$ pixels), which were standardized
by cropping to $3,840\times2,160$ pixels. All images were then tiled
into $640\times360$-pixel tiles using the PILLOW library in Python
(Fig. \ref{fig:Fig4_SegAugZoom}.(a)-(b)). This tiling
strategy enabled the model to train on smaller, localized regions,
improving computational efficiency and reducing the likelihood of
convergence to suboptimal local minima. Following preprocessing, each
tile was manually classified as defect or no defect, forming the final
training dataset. All imagery is of modified-bitumen
	('mod-bit') membranes which is a widely used flat-roof waterproofing
	system comprised of a polymer-modified (APP or SBS) bituminous sheet,
	typically surfaced with mineral granules. This system type was decided
	upon because in-house roofing expertise was available to support reliable
	annotation. Labelling was carried out by one annotator with five years
	of experience as an administrator at a roofing company, who consulted
	in-house roofing experts on ambiguous cases. A tile was labelled defect
	if it contained any recognized mod-bit defect such as surface cracking,
	alligatoring, granular loss, blistering or wrinkling, exposed roof
	deck, moisture staining or a failed seam, or membrane seam separation,
	with representative examples of each shown in Fig. \ref{fig:defectCatalog}.
	Otherwise, the image was labelled as no defect. Since this is a binary
	classifier, no multi-class adjudication was required. After an initial
	pass, the annotator manually reviewed the entire dataset for quality
	control. Because a single expert annotator produced the labels, inter-annotator
	agreement could not be computed, which is noted as a limitation. The
	dataset and model are therefore specific to mod-bit membranes and
	generalization to other systems such as TPO, EPDM, or PVC is not claimed.
The addition of purpose-collected UAS imagery significantly improved
dataset uniformity, coverage of defect types, and the overall fidelity
and robustness of the resulting model.

\subsection{Data augmentation}

The preliminary dataset contained membrane roof images captured from
a variety of angles; however, to expand its diversity and improve
model generalization, additional preprocessing and data augmentation
were applied. Augmentation was performed using the PILLOW library
in Python, employing standard techniques widely used in related research
\textcolor{black}{\cite{Guan2019,Wang2008,Shorten2019}}, including random rotation,
flipping, brightness variation, and noise injection (Fig. \ref{fig:Fig4_SegAugZoom}.(c)-(d)).
Each original image was paired with an augmented counterpart, both
retaining the same class label. These augmentations effectively increased
the dataset size and broadened the range of environmental and lighting
conditions represented. As discussed in Subsection \ref{ssec:image_tiling},
early data sources were limited and relatively homogeneous, which
created classification gaps when the model encountered new UAS-captured
images with different illumination, membrane colours, or surface conditions.
The expanded dataset mitigated these gaps, enabling the model to more
reliably identify defects under previously unseen conditions. Consequently,
the model demonstrated improved robustness and a higher capacity to
generalize beyond the narrow conditions represented in the initial
training set.

\subsection{Image Information Density and Distance\label{ssec:ImageInformationDensity}}

The preliminary dataset consisted of roof images captured at varying
resolutions and distances from the surface. Most online-sourced images
were at least $1,920\times1,080$ pixels, although some lower-resolution
images (down to $1,280\times720$ pixels) were incorporated to broaden
the range of defect conditions represented in the dataset. Image capture
distances by the DJI Matrice 350 RTK drone also varied substantially:
some photographs were taken only one to two meters above the roof
surface, while others were captured from heights of $10$-$20$ meters.
High-resolution, close-range imagery is essential for identifying
fine-grained defects such as membrane cracking or alligatoring, where
subtle texture variations must be clearly visible. Conversely, large-scale
issues, such as missing membrane sections or water pooling are more
effectively assessed using images captured from higher altitudes,
which provide the necessary spatial context. To support the model's
ability to recognize both fine-scale and large-scale defects, images
across this full range of resolutions and distances were retained,
thereby exposing the model to varying levels of effective information
density. Fig. \ref{fig:Fig4_SegAugZoom}.(e) and \ref{fig:Fig4_SegAugZoom}.(f)
illustrate this contrast: the high-resolution source image in \ref{fig:Fig4_SegAugZoom}.(e)
preserves detailed membrane features, whereas the lower-resolution
image in \ref{fig:Fig4_SegAugZoom}.(f) emphasizes broader defect
patterns. For practical UAS data collection, this suggests the value
of a dual-pass strategy: an initial high-altitude survey to capture
large-area anomalies, followed by a low-altitude pass to acquire detailed
imagery for fine-scale defect detection. 
\begin{figure*}[!t]
	\centering{}\includegraphics[scale=0.34]{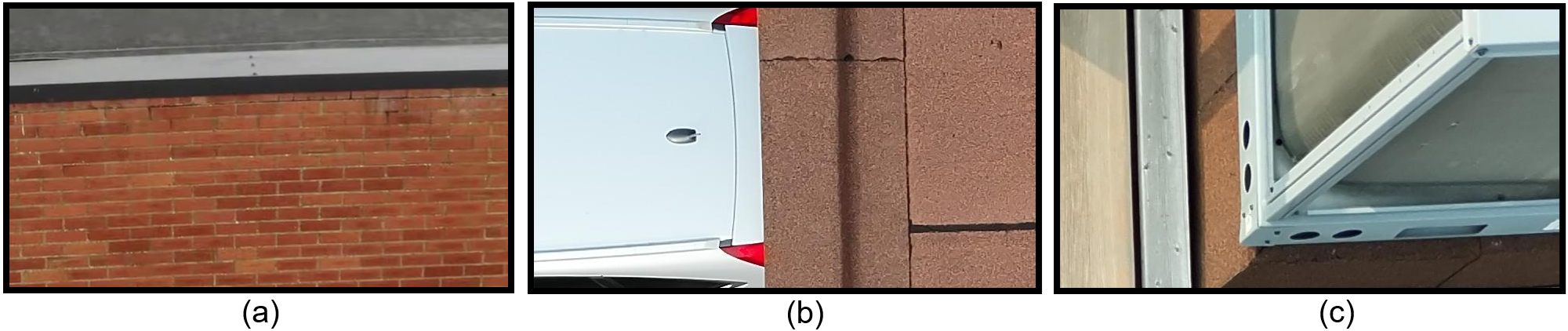}\caption{\label{fig:Fig5_DatasetExclusion}Examples of images that have been
		excluded from the dataset: (a) an image predominantly displaying a
		brick wall with some membrane and flashing visible, (b) the edge of
		a flat roof, with a vehicle visible on one side below the roof, (c)
		a flat roof with a wooden deck on top on the left side, as well as
		a mechanical unit on the right side.}
\end{figure*}

\subsection{Dataset Compilation and Exclusions}

During dataset preparation, specific image types were intentionally
excluded to facilitate more effective model training. The goal was
to provide the algorithm with clean, idealized examples during the
early stages of development by removing images containing elements
that could introduce noise or confuse defect classification. Representative
examples of excluded images are shown in Fig. \ref{fig:Fig5_DatasetExclusion}.
These images contain objects commonly present on rooftops or within
the camera's field of view but unrelated to roof condition assessment.
For instance, Fig. \ref{fig:Fig5_DatasetExclusion}.(a) includes a
brick wall that does not contribute any information about the state
of the roof and may distract the model. Similarly,
Fig. \ref{fig:Fig5_DatasetExclusion}.(b) captures the roof edge along
with a vehicle in an adjacent parking lot, introducing irrelevant
visual features that could negatively affect training. In Fig. \ref{fig:Fig5_DatasetExclusion}.(c),
portions of a wooden plank and a mechanical unit appear alongside
a small section of membrane, a scenario frequently encountered during
UAS roof inspections but not informative for early-stage defect detection.
These examples represent only a subset of the images that were removed;
the full set is non-exhaustive but illustrates the rationale behind
the exclusion process. As the dataset expands and the model matures,
the long-term objective is to reintroduce such images properly annotated
to improve robustness and generalization. However, to avoid early-stage
overfitting and reduce classification ambiguity, these images were
omitted from the initial phases of this research.

\section{Methodology\label{sec:methodology}}

CNNs are widely recognized as the primary tool for modern image classification
due to their robustness and adaptability across diverse applications
\textcolor{black}{\cite{Kim2017,Bouguettaya2022,Krizhevsky2012,Tang2015,Aditya2024_SegNet}}.
CNNs employ multiple filters to extract hierarchical features from
images, enabling effective classification \textcolor{black}{\cite{Kim2017,Bouguettaya2022,Aditya2024_SegNet}}.
In this research, a CNN is applied to classify roof surfaces as defective
or non-defective using live imagery collected by a
UAS. Roof images are largely static aside from camera movement, reducing
complications from dynamic environmental changes. To achieve near
real-time processing, this study incorporates image tiling
(Section \ref{ssec:image_tiling}), which simplifies the classification
task. Images are composed of pixels representing the underlying information,
which can be interpreted as features such as color, texture, shape,
and spatial location, analogous to human visual perception. CNNs extract
these features through convolutions, where filters combine input information
to produce feature maps. Early convolutional layers detect simple
patterns, such as edges or corners, while deeper layers generate complex
feature maps representing higher-level structures. The resolution
of the input image directly affects the feature map's detail: higher-resolution
images yield richer feature maps, enhancing the model's ability to
detect fine defects. The architecture of a CNN, including the number,
order, and size of filters, as well as pooling operations, is application-specific.
Pooling reduces the dimensionality of feature maps by aggregating
neighboring information, balancing computational efficiency with feature
preservation. For roof defect detection, a higher number of feature
maps is necessary to capture the diversity of potential defects. This
necessitates a large, varied dataset to avoid underfitting, as discussed
in Section \ref{sec:problemformulation}. However, excessive augmentation
or tiling without sufficient data diversity can
lead to overfitting, highlighting the need for careful dataset design.
Standard practice in image classification often uses square images
of $256\times256$ pixels to reduce computational demand \cite{Krizhevsky2012}.
While convenient, such downscaling compromises feature map resolution
and is insufficient for real-world UAS imagery, where higher-resolution
inputs are required to detect both small-scale defects, like membrane
cracking, and larger-scale issues, such as pooling water or missing
membranes. 
\begin{figure*}[!t]
	\centering{}\includegraphics[scale=0.19]{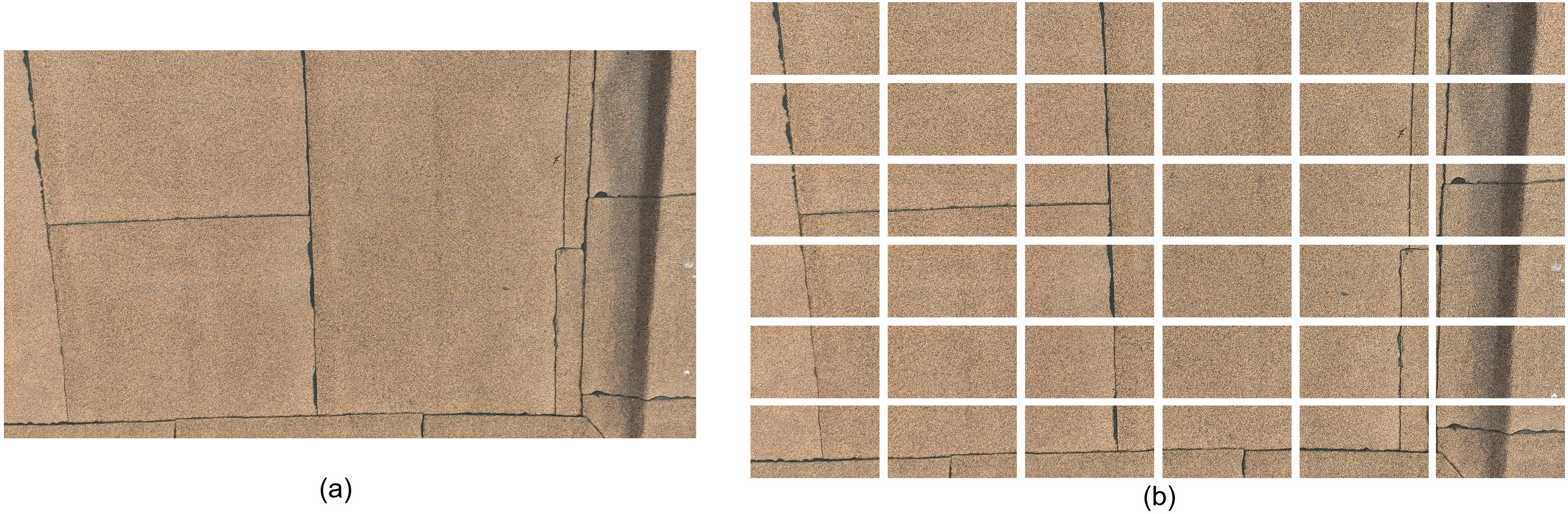}\caption{\label{fig:imageSegmentation}tiling approach: (a)
		An image captured using a UAS and (b) the resulting tiling
		of the same image into 36 equal parts. The tiling
		of the image allows for faster compute time while maintaining all
		of the information that is important to the classification problem.}
\end{figure*}

\subsection{tiling}

\begin{figure*}[!t]
	\centering{}\includegraphics[width=0.9\textwidth]{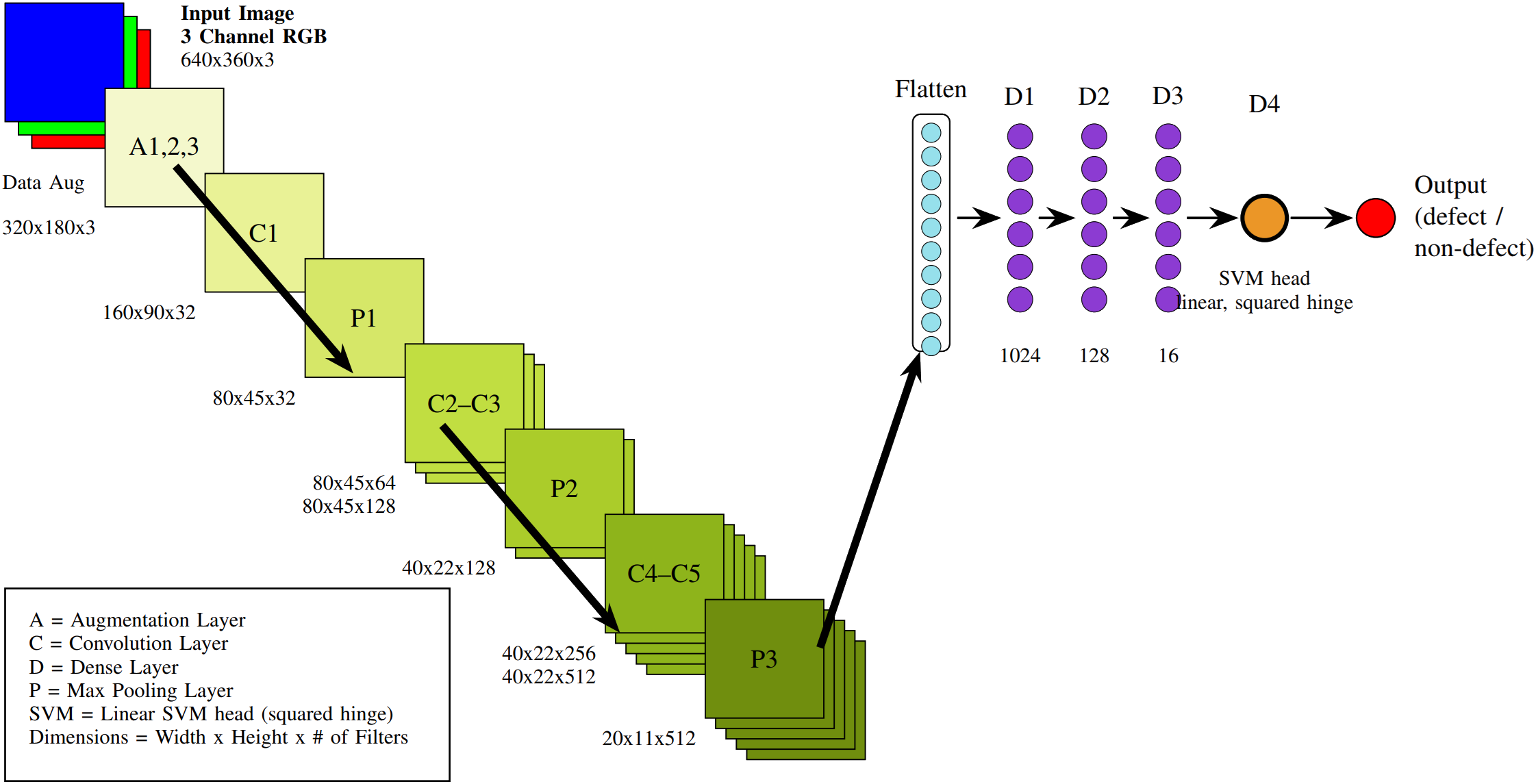}
	\caption{\label{fig:Fig7_ModelArchitect}Proposed model architecture: following
		the image tiling process, RGB images are input to
		the model, and are processed through data augmentation, convolution,
		and pooling layers before being flattened. From here, the data is
		processed through four dense layers, \textcolor{black}{the last of which is a linear squared-hinge Support Vector Machine (SVM) head that produces the binary defect~/~non-defect decision}. The
		exact specifications of each layer used can be seen in Table \ref{tab:ModelArchitecture}.}
\end{figure*}

To maintain a real-world data stream while meeting real-time processing
requirements, the collected images are pre-processed through image
tiling prior to input into the CNN. This approach
reduces feature map sizes, thereby decreasing computational load and
training times, while still preserving the high-resolution detail
necessary for accurate defect classification. As described in Section
\ref{ssec:datasetPreparation}, the initial training data were sourced
from online video repositories, whereas subsequent testing and additional
training data were acquired using a UAS (DJI Matrice 350 RTK drone
via onsite visits). Image tiling divides a high-resolution
image into smaller, lower-resolution components. These tiles
collectively retain the total information content and pixel count
of the original image, and can be recombined if needed. However, larger
objects spanning multiple tiles may lose their overall
form, which can complicate learning or classification. This limitation
is mitigated by capturing images at varying heights relative to the
roof surface, as discussed in Section \ref{ssec:ImageInformationDensity}.
Objects that lose form at lower altitudes can be better interpreted
through images acquired at higher altitudes. While this tiling
strategy is particularly suitable for rooftop defect detection, its
applicability to other classification tasks may be limited. Defects
often vary in shape, size, and color, and may span across multiple
tiles \textcolor{black}{\cite{Gullbrekken2016,Carretero-Ayuso2016,Zahradnik2024}},
allowing classification to proceed effectively. The number of tiles
generated from each original image is determined by its resolution.
A final tile size of $640\times360$ pixels was
selected, allowing all source images to be evenly divided while facilitating
manual labeling and reducing processing times. Images of $1,280\times720$
pixels were divided into four tiles, $1,920\times1,080$
pixel images into nine tiles, and high-resolution
UAS imagery of $3,840\times2,160$ pixels into $36$ tiles.
This systematic tiling ensures consistent input
dimensions for the model while retaining sufficient detail for accurate
defect detection.
\[
\frac{\text{Width of Cropped Image}}{\text{Number of Vertical Sectors}}=\frac{3840}{6}=640\text{ pixels}
\]
\[
\frac{\text{Height of Cropped Image}}{\text{Number of Horizontal Sectors}}=\frac{2160}{6}=360\text{ pixels}
\]
The use of tiled images provides several key advantages.
Primarily, defects are often visible in only a single tile,
enabling more precise classification and thereby improving training
accuracy. Additionally, smaller tiled images contain
fewer pixels, resulting in reduced feature map sizes within the CNN
and significantly shorter training and processing times compared to
networks processing full-resolution images. A key objective of this
project is to integrate the classification model with a UAS. The UAS
will capture sequential, overlapping images of the roof surface at
a resolution of $3,840\times2,160$ pixels and process them in real
time to identify defects. Photos, rather than video, are used to simplify
the processing pipeline. High frame rates are unnecessary for two
reasons: first, the substantial overlap between images ensures that
any defect visible in one image will also appear in neighboring images;
second, the roof surface is static, making processing speed more critical
than capture rate. The current model architecture is shown in Fig.
\ref{fig:Fig7_ModelArchitect}. Experiments with an $80\%$ image
overlap yielded successful results at capture rates of $0.5$-$1.4$
images per second, which substantially reduces computational demands
relative to continuous video capture. While video has been used for
training purposes, slower capture rates are sufficient for real-time
classification. Limiting the data input rate allows the UAS to perform
additional tasks concurrently, such as generating a live map of identified
defects for the operator, while maintaining real-time performance.

\begin{figure*}[!ht]
	\centering{}\includegraphics[width=1\textwidth]{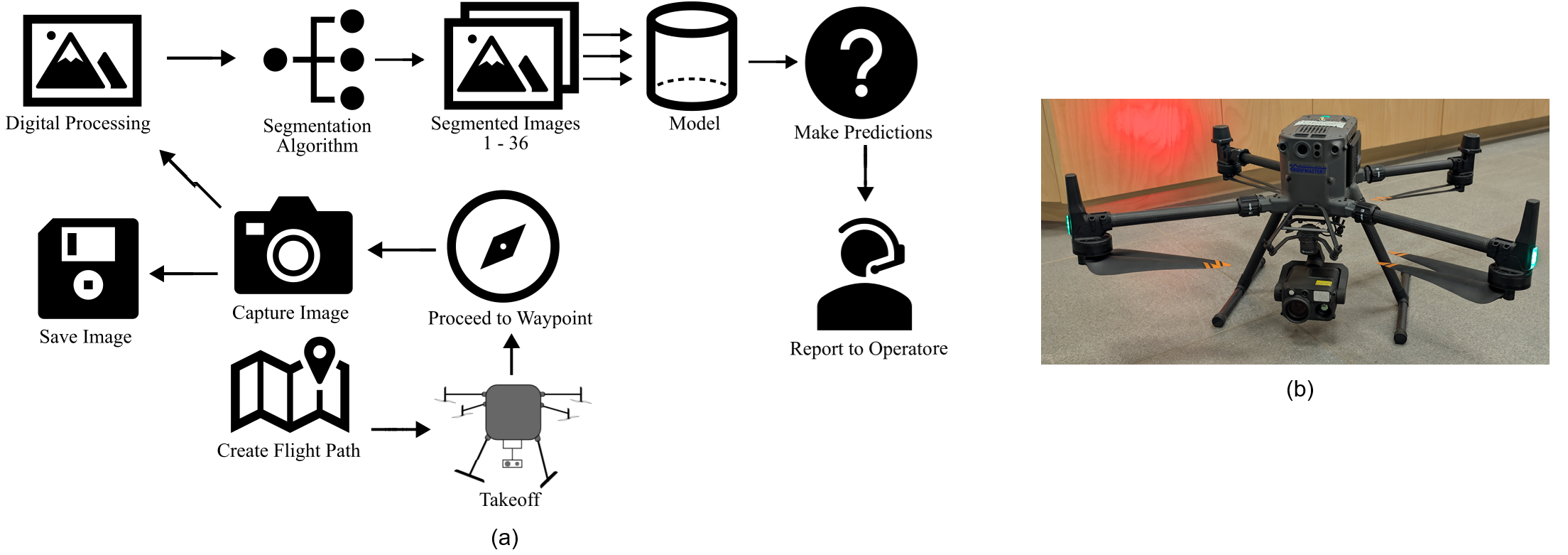}
	\caption{\label{fig:Fig8_uasWorkflow}Experimental Validation: (a) Proposed
		UAS workflow for autonomous rooftop defect detection, and (b) DJI
		Matrice 350 RTK UAS utilized to collect training and testing data.}
\end{figure*}

\subsection{CNN-SVM Model architecture}

The CNN employed here is coupled with an SVM, a
method widely used in similar studies for binary classification \cite{Agarap2019}.
SVMs are supervised learning models designed primarily for classification
tasks \textcolor{black}{\cite{Cortes1995,Tang2015}}. Their objective is to identify
an optimal hyperplane that distinctly separates data points of different
classes in a high-dimensional feature space. This approach has demonstrated
robust performance across diverse applications. The activation function
used in this model is the Rectified Linear Unit (ReLU), which facilitates
learning of complex patterns while maintaining computational efficiency.
For linearly separable data, the SVM algorithm aims to maximize the
margin of the separating hyperplane. The optimization problem is mathematically
expressed as in Eq. \eqref{eq:svm1}, \eqref{eq:svm2}, and \eqref{eq:svm3}.
In Eq. \eqref{eq:svm1}, $w$ represents the weights, $b$ the bias,
and
\begin{equation}
	f(w,x)=w^{\top}x+b\label{eq:svm1}
\end{equation}
The equation
\begin{equation}
	\min\frac{1}{2}\|w\|_{2}^{2}+C\sum_{i=1}^{p}\max\left(0,\,1-y_{i}(w^{\top}x_{i}+b)\right)\label{eq:svm2}
\end{equation}
where $p$ is the number of training points, $\|w\|_{2}^{2}=w^{\top}w$
	is the squared $\mathcal{L}_{2}$ norm (the margin regulariser), $C$
	is the penalty parameter, $y_{i}\in\{-1,+1\}$ is the label of the
	$i$-th point, and $w^{\top}x_{i}+b$ is its decision value. Equation~\eqref{eq:svm2}
	uses the hinge loss; replacing it with the squared hinge gives the
	differentiable $\mathcal{L}_{2}$-SVM adopted here,
\begin{equation}
	\min\frac{1}{2}\|w\|_{2}^{2}+C\sum_{i=1}^{p}\max\left(0,\,1-y_{i}(w^{\top}x_{i}+b)\right)^{2}\label{eq:svm3}
\end{equation}
which penalises margin violations more strongly and
	is differentiable, easing gradient-based optimisation; this squared-hinge
	output replaces the conventional sigmoid cross-entropy head used in
	standard CNN classifiers. \textcolor{black}{At inference, each tile is 
	assigned to the defective class when the SVM decision value $w^{\top}x+b\geq0$,
	and to the non-defective class otherwise. This corresponds with a fixed 
	decision threshold of zero.} The final CNN model consists of five convolutional
layers interspersed with max pooling layers. Following these layers,
the output is flattened and passed through four dense layers\textcolor{black}{, the last being a linear squared-hinge SVM head}. The
detailed layer architecture is provided in Table \ref{tab:ModelArchitecture}.
\begin{table}[!ht]
	\centering{}\caption{\label{tab:ModelArchitecture}Model Architecture.}
	\begin{tabular}{c>{\raggedright}p{7.4cm}}
		\toprule 
		Layer  & Description\tabularnewline
		\midrule
		\midrule 
		1  & Input layer: Input Shape = (360, 640, 3)\tabularnewline
		\midrule 
		2  & Resizing layer: Output Shape = (180, 320)\tabularnewline
		\midrule 
		3  & Data Augmentation Layer: Random Rotation = 0.1\tabularnewline
		\midrule 
		4  & Data Augmentation Layer: Random Brightness = 0.1\tabularnewline
		\midrule 
		5  & 1st Convolution 2D layer: Filters = 32, Activation function = ReLu,
		Kernel size = 3, Stride = 2, Padding = ``same''\tabularnewline
		\midrule 
		6  & 1st Max Pool 2D Layer: Pool size = 2, Stride = 2\tabularnewline
		\midrule 
		7  & 2nd Convolution 2D layer: Filters = 64, Activation function = ReLu,
		Kernel size = 3, Padding = ``same''\tabularnewline
		\midrule 
		8  & 3rd Convolution 2D layer: Filters = 128, Activation function = ReLu,
		Kernel size = 3, Padding = ``same''\tabularnewline
		\midrule 
		9  & 2nd Max Pool 2D layer: Pool size = 2, Stride = 2\tabularnewline
		\midrule 
		10  & 4th Convolution 2D layer: Filters = 256, Activation function = ReLu,
		Kernel size = 3, Padding = ``same''\tabularnewline
		\midrule 
		11  & 5th Convolution 2D layer: Filters = 512, Activation function = ReLu,
		Kernel size = 3, Padding = ``same''\tabularnewline
		\midrule 
		12  & 3rd Max Pool 2D layer: Pool size = 2, Stride = 2\tabularnewline
		\midrule 
		13  & Flatten layer\tabularnewline
		\midrule 
		14  & 1st Dense layer: Units = 1024, Activation function = ReLu, Regularizer
		= L2, L2 = 0.01\tabularnewline
		\midrule 
		15  & 2nd Dense layer: Units = 128, Activation function = ReLu, Regularizer
		= L2, L2 = 0.01\tabularnewline
		\midrule 
		16  & 3rd Dense layer: Units = 16, Activation function = ReLu, Regularizer
		= L2, L2 = 0.01\tabularnewline
		\midrule 
		17  & \textcolor{black}{4th Dense layer (linear SVM head): Units = 1, Activation Function = Linear, Loss = squared hinge ($\mathcal{L}_2$-SVM)}\tabularnewline
		\bottomrule
	\end{tabular}
\end{table}

\section{Workflow, Challenges, and Mitigation\label{sec:WorkflowChalleng}}

\subsection{Overfitting and Optimization Issues\label{ssec:Overfitting}}

Preliminary model development revealed a persistent performance issue:
the accuracy and loss metrics plateaued below $70\%$ during training.
This behavior was attributed to overfitting. Although data augmentation
slightly improved performance, it did not resolve the problem. An
early-stopping callback was introduced to halt training when the loss
failed to decrease; however, the callback triggered immediately and
repeatedly, indicating that the model was not converging effectively.
In the final implementation, the callback was configured to monitor
loss values with a minimum improvement threshold of $0.001$. The
primary resolution emerged from modifying the optimizer. Early development
relied on the Adam optimizer, which combines momentum and root-mean-square
propagation for adaptive learning. However, replacing Adam with Stochastic
Gradient Descent (SGD) \cite{Poojary2019} produced a substantial
improvement. Unlike Adam, SGD uses a fixed learning rate; a low learning
rate of $0.005$ and a momentum value of $0.2$ were applied. After
switching to SGD, model accuracy increased markedly, exceeding $90\%$.
Additional adjustments were made to refine model performance. The
early-stopping callback, which previously halted training around the
third epoch, no longer triggered prematurely once the optimizer was
corrected. Following the initial architectural configuration, the
model underwent an extended training cycle before new test data were
introduced. This test dataset collected using the UAS workflow specifically
for this research. Subsequent evaluation by additional previously
unseen test dataset produced accuracy values up to $93.2\%$, aligning
with the expected performance.

\subsection{Workflow}

\begin{figure*}[!t]
	\centering{}\begin{tikzpicture}[x=0.72cm,y=0.72cm]
		\foreach \x in {0,...,5} {\foreach \y in {0,...,5} {\draw[fill=green!35,draw=gray] (\x,\y) rectangle ++(1,1);}}
		\foreach \rx/\ry in {2/4,3/4,1/1,4/2} {\draw[fill=red!55,draw=gray] (\rx,\ry) rectangle ++(1,1);}
		\node[anchor=south,font=\small\bfseries] at (3,6.15) {(a) Per-tile prediction grid};
		\node[anchor=north,font=\scriptsize] at (3,-0.15) {one UAS image, $6\times6$ tiles};
		\draw[fill=green!35,draw=gray] (0,-1.5) rectangle ++(0.55,0.55); \node[anchor=west,font=\scriptsize] at (0.6,-1.22) {no defect};
		\draw[fill=red!55,draw=gray] (2.7,-1.5) rectangle ++(0.55,0.55); \node[anchor=west,font=\scriptsize] at (3.3,-1.22) {defect};
	\end{tikzpicture} \hspace{1.2cm} \begin{tikzpicture}[x=0.72cm,y=0.72cm]
		\draw[thick,fill=gray!12] (0,0) rectangle (6,6);
		\node[anchor=south,font=\small\bfseries] at (3,6.15) {(b) Live defect map};
		\node[anchor=north,font=\scriptsize] at (3,-0.15) {aggregated along the flight path};
		\draw[dashed,gray] (0.6,5.4) -- (5.4,5.4) -- (5.4,3.9) -- (0.6,3.9) -- (0.6,2.4) -- (5.4,2.4) -- (5.4,0.9) -- (0.6,0.9);
		\foreach \mx/\my in {1.6/5.3,4.7/4.0,2.3/2.5,3.6/1.0} {\fill[red] (\mx,\my) circle (3.2pt);}
	\end{tikzpicture} \caption{\label{fig:realtime}Illustrative (mock-up) operator
			feedback during a flight. (a) For each captured UAS image, the model
			classifies its $36$ tiles and presents a colour-coded grid (green:
			no defect, red: defect). (b) Detected defects are aggregated into
			a live map along the flight path for operator review and revisit-waypoint
			generation.}
\end{figure*}
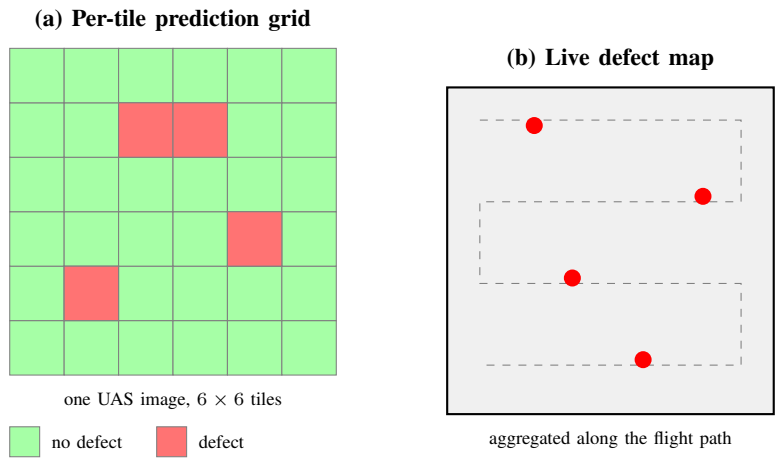

The CNN-SVM model developed in this research is
intended to support a key stage of the workflow illustrated in Fig.
\ref{fig:Fig8_uasWorkflow}. The design philosophy throughout development
was guided by the requirements for eventual deployment on a UAS capable
of near real-time onboard processing. During operation, the UAS (DJI
Matrice 350 RTK drone) is programmed to follow a predefined flight
path to scan the roof surface. Two passes are conducted at different
altitudes to provide complementary levels of detail: higher-altitude
imagery captures large-scale structural issues, while lower-altitude
imagery resolves fine-grained defects. This multi-scale approach is
further discussed in Section \ref{ssec:ImageInformationDensity}.
During each pass, the UAS captures images with an overlap of approximately
$50\%$-$90\%$ at a rate of $1$-$2$ images per second. Immediately
after capture, each image is temporarily stored, creating a buffer
that facilitates continuous operation while maintaining real-time
processing. A short delay between capture and classification is acceptable
given the static nature of roof surfaces and allows the UAS to perform
additional critical tasks as required. Once stored, each image is
tiled into the appropriate sub-images before being
passed to the classification model. The tiled images
are processed sequentially at a resolution of $640\times360$ pixels.
The proposed CNN-SVM model, designed to be computationally
lightweight and accurate, employs a moderate number
of layers to balance computation constraints with classification performance.
The output of the CNN-SVM model is a binary prediction
indicating whether each tile contains a defect.
After all tiles from an image have been evaluated,
the predictions are presented to the operator in a grid format (illustrated
	conceptually in Fig.~\ref{fig:realtime}).

\vspace{0.2cm}
With an image capture rate of approximately one frame per second,
this workflow provides operators sufficient time to assess detected
defects and determine whether the scanned region is clear. At the
conclusion of the roof inspection, all results are compiled into a
report. If defects are detected, the system can initiate secondary
actions, such as generating additional waypoints for revisiting specific
locations to conduct further analysis. To avoid overstating
	the system's maturity, it is necessary to distinguish what is implemented
	and evaluated in this work from what is proposed for future development.
	Firstly, the tile-level CNN-SVM classifier, the training and evaluation
	pipeline, and all reported metrics, and inference-time results have
	been implemented and evaluated. Future work that has not yet been
	implemented includes closed-loop autonomous flight with automatic
	waypoint generation, on-board execution, georeferenced defect mapping,
	and automated revisiting of flagged regions. The workflow described
	above therefore represents the intended end-to-end concept, of which
	only the per-tile classification stage is realized and benchmarked
	in this paper; the remaining stages are directions for future work.

\begin{table*}[!t]
	\centering{}\caption{\label{tab:ModelPerformance}Architecture sweep of
			the proposed CNN-SVM on the dataset: held-out test accuracy, training
			accuracy, training time, and epochs for $2$--$6$ convolutional
			and $2$--$4$ dense layers (single seed, $100$-epoch cap, identical
			hyperparameters; only the layer counts vary). The adopted $5$-convolution/$4$-dense
			configuration is shown in bold.}
	\begin{tabular}{llccccc}
		\toprule 
		&  & 2 Conv  & 3 Conv  & 4 Conv  & 5 Conv  & 6 Conv\tabularnewline
		\midrule
		\midrule 
		\multirow{4}{*}{2 Dense} & Test acc.  & 92.5\%  & 87.8\%  & 92.0\%  & 91.7\%  & 94.5\%\tabularnewline
		& Train acc.  & 92.7\%  & 93.3\%  & 94.4\%  & 94.6\%  & 96.6\%\tabularnewline
		& Time  & 2h 20m  & 1h 22m  & 2h 24m  & 1h 44m  & 2h 48m\tabularnewline
		& Epochs  & 100  & 100  & 100  & 100  & 100\tabularnewline
		\midrule 
		\multirow{4}{*}{3 Dense} & Test acc.  & 89.8\%  & 92.1\%  & 91.1\%  & 93.5\%  & 94.1\%\tabularnewline
		& Train acc.  & 94.3\%  & 95.6\%  & 96.1\%  & 96.5\%  & 96.7\%\tabularnewline
		& Time  & 2h 16m  & 1h 21m  & 2h 25m  & 1h 44m  & 2h 53m\tabularnewline
		& Epochs  & 100  & 100  & 100  & 100  & 100\tabularnewline
		\midrule 
		\multirow{4}{*}{4 Dense} & Test acc.  & 90.0\%  & 91.3\%  & 92.3\%  & \textcolor{red}{\textbf{94.1\%}}  & 94.3\%\tabularnewline
		& Train acc.  & 94.1\%  & 95.4\%  & 96.0\%  & \textcolor{red}{\textbf{96.3\%}}  & 96.6\%\tabularnewline
		& Time  & 2h 30m  & 1h 24m  & 2h 40m  & \textcolor{red}{\textbf{1h 45m}}  & 2h 52m\tabularnewline
		& Epochs  & 100  & 100  & 100  & \textcolor{red}{\textbf{100}}  & 100\tabularnewline
		\bottomrule
	\end{tabular}
\end{table*}

\begin{table}[!ht]
	\caption{\label{tab:ImagingBreakdown}Image and classification breakdown.}
	
	\centering{}%
	\begin{tabular}{ccccc}
		\toprule 
		\multicolumn{5}{c}{Data Categorization}\tabularnewline
		\midrule 
		Images  & Original Full  & tiled  & Augmented  & Total\tabularnewline
		\midrule
		\midrule 
		Defect  & -  & 11,057  & 8,874  & 19,931\tabularnewline
		\midrule 
		No Defect  & -  & 14,691  & 15,170  & 29,861\tabularnewline
		\midrule 
		Total  & 5,240  & 25,748  & 24,044  & 49,792\tabularnewline
		\bottomrule
	\end{tabular}
\end{table}

\begin{table}[!ht]
	\centering{}\caption{\label{tab:TrainingConsideration}Effect of the two
			training choices that most affect the proposed CNN-SVM, on the dataset
			(single seed, $100$-epoch cap). Replacing SGD with Adam collapses
			training to the majority class; removing data augmentation fits the
			training set almost perfectly and widens the train-test gap. These
			single-run values carry run-to-run variation; the seed-averaged accuracy
			is reported in Table~\ref{tab:modelComparisons}.}
	\begin{tabular}{lccc}
		\toprule 
		Configuration  & Train acc.  & Test acc.  & Recall\tabularnewline
		\midrule
		\midrule 
		Proposed (SGD\,+\,augmentation)  & $96.4\%$  & $91.4\%$  & $0.940$\tabularnewline
		\midrule 
		Without augmentation & $99.7\%$  & $92.3\%$  & $0.875$\tabularnewline
		\midrule 
		Adam instead of SGD  & $49.5\%$ & $54.4\%$  & $0.000$\tabularnewline
		\bottomrule
	\end{tabular}
\end{table}

\section{Experimental Results\label{sec:results}}

The UAS utilized to collect training and testing data for this project
was a DJI Matrice 350 RTK. The camera mounted to this UAS was a DJI
Zenmuse H30T. The M350 weighs in at 6.47 kg with
batteries and before payload, with a maximum flight time of 55 minutes.
The M350 RTK connects with GNSS networks to maintain hovering accuracy,
and has a maximum transmission distance of 20 km. The H30T payload
weighs 920 g and provides 3-axis stabilization in
flight. When connected with the UAS the camera has a controllable
pan range of $320^{\circ}$and a tilt range of $-120^{\circ}$ to
$+60^{\circ}$. The H30T has three cameras (wide-angle, zoom, and
infrared), as well as a laser range-finder. The wide-angle camera
used to collect imagery for this research uses a 1/1.3-inch CMOS sensor,
has an Aperture of f/1.7 and has an effective resolution of 48 MP.

\subsection{Dataset Download}

To ensure reproducibility, all dataset splits, core
	scripts, and trained models that we collected, processed, and produced
	are openly archived on Zenodo under a Creative Commons Attribution
	4.0 International (CC BY 4.0) license. A fixed random seed of 42 was
	maintained across all experiments to guarantee deterministic results.
	The specific assets are available via the following Digital Object
	Identifiers (DOIs):
\begin{itemize}
	\item The 'tiled and augmented UAS imagery of modified-bitumen
		flat-roof defects' dataset, including the reproducible train/validation/test
		split index (manifest.csv), is accessible at \href{https://doi.org/10.5281/zenodo.20594605}{https://doi.org/10.5281/zenodo.20594605}
	\item The scripts utilized for tiling, augmentation, dataset
		splitting, training, and evaluation are archived at \href{https://doi.org/10.5281/zenodo.20599364}{https://doi.org/10.5281/zenodo.20599364}.
	\item The final trained model weights are preserved at
	\href{https://doi.org/10.5281/zenodo.20596485}{https://doi.org/10.5281/zenodo.20596485}.
\end{itemize}

\subsection{Proposed CNN-SVM Performance and Test Accuracy}

The CNN-SVM model developed in this research was
trained on a lab computer equipped with an AMD Ryzen 9800X3D CPU (4.7
GHz, 4 nm) and an Nvidia RTX 5070 GPU (192-bit memory interface, 2.33
GHz). Python 3.13 was used, along with TensorFlow, Keras, NumPy, Matplotlib,
and PILLOW. Model iterations were performed incrementally, with small
hyperparameter adjustments to ensure controlled and traceable improvements.
A video summarizing the final autonomous CNN-SVM
framework for drone-based rooftop defect inspection is available at
\begin{itemize}
	\item Video of experiment: \href{https://youtu.be/9RsdTeuyIJM}{https://youtu.be/9RsdTeuyIJM}
\end{itemize}
Table~\ref{tab:ModelPerformance} reports the training
	and test accuracy, training time, and number of epochs for the tiled-image
	model across $2$--$6$ convolutional and $2$--$4$ dense layers.
	This testing was evaluated with identical callback and optimizer settings
	and a $100$-epoch cap (the callback configuration is detailed in
	Section~\ref{ssec:Overfitting}). On this split none of the configurations
	degenerates: every architecture attains between $87.8\%$ and $94.5\%$
	test accuracy. Increasing the number of convolutional layers yields
	a small but consistent improvement (from $\approx\!90.8\%$ mean test
	accuracy at two convolutional layers to $\approx\!94.3\%$ at six,
	averaged over dense depth). Conversely, the number of dense layers
	has little effect ($\approx\!91.7\%$ to $92.4\%$ across two to four).
	Training accuracy rises monotonically with depth (up to $96.7\%$),
	but test accuracy plateaus, indicating that the additional capacity
	mainly improves the fit to the training data rather than generalization.
	The adopted five-convolution/four-dense model reaches $94.1\%$ test
	accuracy, within roughly half a percentage point of the best configuration
	(six convolutional layers, $94.3$--$94.5\%$) however it's training
	time was about an hour faster. It was therefore retained as a balance
	between accuracy and computational cost. Since each cell reflects
	a single training run, differences below roughly two percentage points
	are within run-to-run variation (see the seed-averaged confidence
	intervals in Table~\ref{tab:modelComparisons}). Throughout development,
key performance metrics were continuously monitored to ensure sustained
improvement in both training and validation accuracy. Table \ref{tab:ImagingBreakdown}
shows the number of original images and their tiling
into the dataset used for training. After tiling,
some images were removed for quality control, and data augmentation
was applied to increase dataset diversity.

\vspace{0.2cm}
The dataset was divided at the level of whole source
	photographs into training ($75\%$), validation ($15\%$), and test
	($10\%$) sets using a fixed random seed ($42$). With this, every
	tile and augmentation derived from a given photograph remains within
	a single split. No photograph appears in more than one split. The
	resulting sets contain $43,383$ training, $3,869$ validation, and
	$2,540$ test images; augmentations are retained in the training set
	only, while validation and test use original tiles only. Balanced
	class weights computed from the training set were applied during training
	to counter the mild class imbalance, without resampling or threshold
	calibration. Because the split is at the photograph level, different
	(spatially overlapping) photographs of the same roof can still fall
	in different splits. The residual effect of these overlapping images
	is acknowledged in the Limitations and Future Work subsection. Table~\ref{tab:classDist}
	summarizes the class distribution across the splits.

\begin{table}[!ht]
	\centering{}\caption{\label{tab:classDist}Class distribution across the
			training, validation, and test splits.}
	\begin{tabular}{lccc}
		\toprule 
		Set  & Defect  & No defect  & Total\tabularnewline
		\midrule
		\midrule 
		Training  & 17,185  & 26,198  & 43,383\tabularnewline
		\midrule 
		Validation  & 1,589  & 2,280  & 3,869\tabularnewline
		\midrule 
		Test  & 1,157  & 1,383  & 2,540\tabularnewline
		\bottomrule
	\end{tabular}
\end{table}

Table \ref{tab:TrainingConsideration} isolates the
	two training choices that most affect the proposed model on the dataset.
	The choice of optimizer is clear. When trained with the Adam optimizer
	the network fails to converge, collapsing to the majority class (a
	test accuracy of $54.4\%$ at zero recall). However, when utilizing
	Stochastic Gradient Descent with momentum, the model trains successfully.
	Data augmentation, in turn, curbs overfitting. Without it the model
	fits the training data almost perfectly ($99.7\%$ training accuracy)
	which causes the train-test gap to widen. When trained with augmented
	data, the model generalizes with a smaller gap at a comparable test
	accuracy. These single-run values carry a few percentage points of
	run-to-run variation (see the seed-averaged results in Table~\ref{tab:modelComparisons}).
	The qualitative conclusions from this testing show that SGD is essential
	and augmentation reduces overfitting. Once a stable training setup
	was established, evaluation shifted to the held-out test set. Following
iterative revisions, the proposed CNN-SVM model
achieved approximately $97.6\%$ training accuracy,
	$94.0\%$ validation accuracy, and a mean test accuracy of $94.4\%$
	($95\%$ CI $\pm0.4\%$ over three seeds) (see
Table \ref{tab:modelComparisons}), with a processing
	speed of roughly $130$ full $36$-tile UAS images per second ($\approx7.8$\,ms
	per image). These results demonstrate that the CNN-SVM
model meets the project goals of high classification accuracy and
efficient computation. With further refinement and training, the model
is well positioned for full-scale deployment as an autonomous roof-inspection
solution.

\begin{figure*}[!ht]
	\centering{}\includegraphics[width=1\textwidth]{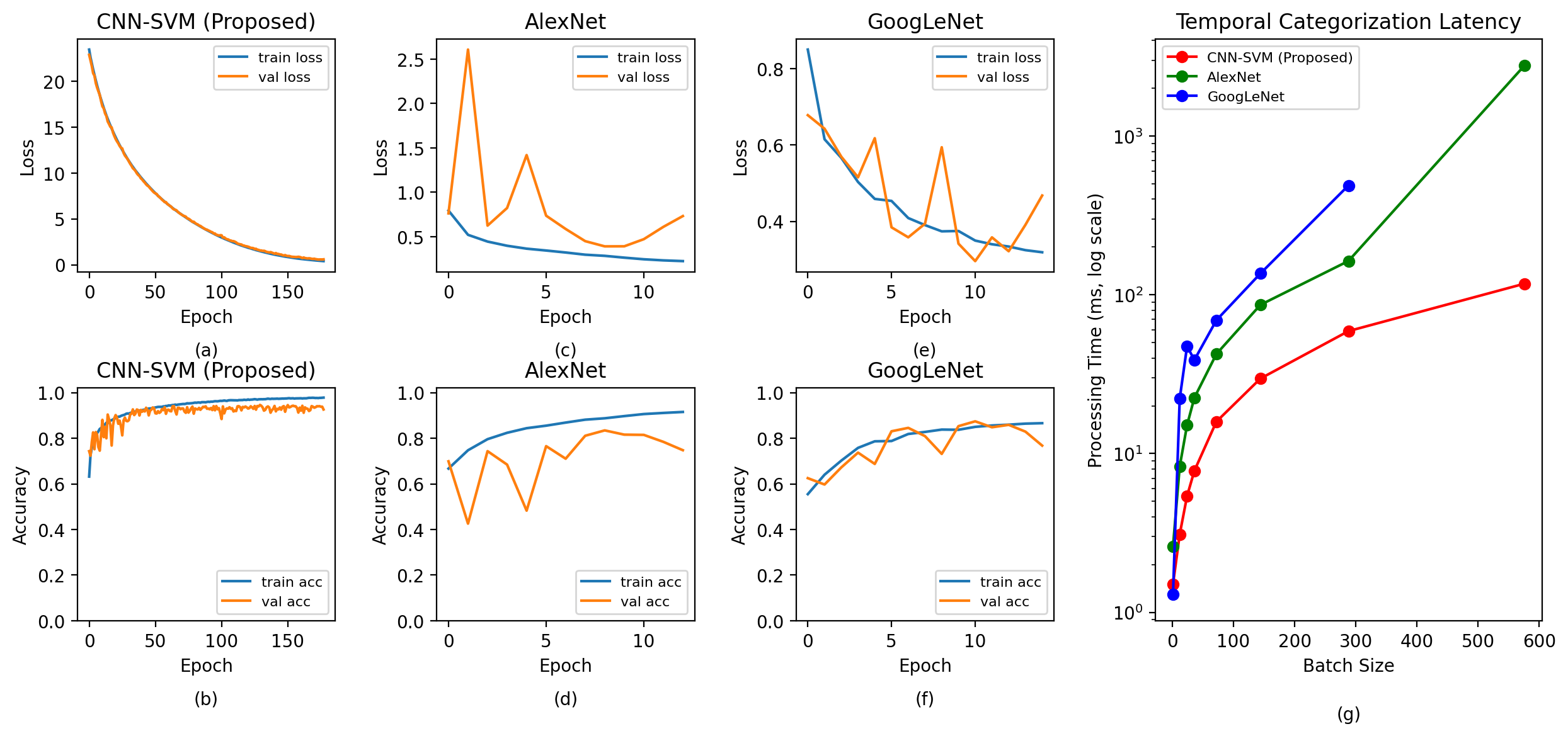}\caption{\label{fig:Fig9_AccLossTempor}The left six subfigures presents accuracy
		((b), (d), and (f)) and losses ((a), (c), and (e)) comparison between
		cutting edge machine learning techniques across both training and
		validation: Proposed CNN-SVM ((a) and (b)), AlexNet
		((c) and (d)), and GoogLeNet ((e) and (f)). The right portion (g)
		refers to Temporal Categorization Latency chart which represents the
		time required to perform classification task with different batch
		sizes of input images. The latency axis in (g) is
			logarithmic; GoogLeNet exceeded GPU memory at the largest batch size
			($576$ tiles) and is not plotted there.}
\end{figure*}

\begin{table*}[!t]
	\centering{}\caption{\label{tab:modelComparisons}Model comparison on
			the test set ($n=2,540$): mean $\pm$ $95\%$ confidence interval
			over three random seeds. GoogLeNet uses seeds $42/43/45$   seed $44$
			diverged to a NaN training loss in the first epoch and was replaced
			(the failed run is retained as an artifact); CNN-SVM and AlexNet use
			seeds $42$--$44$.}
	\begin{tabular}{lccc}
		\toprule 
		Model  & Test Accuracy ($95\%$ CI)  & Recall  & Precision\tabularnewline
		\midrule
		\midrule 
		CNN-SVM (Proposed)  & $\mathbf{94.4\%\pm0.4\%}$  & $0.920\pm0.025$  & $0.956\pm0.030$\tabularnewline
		\midrule 
		GoogLeNet  & $89.2\%\pm5.7\%$  & $0.883\pm0.156$  & $0.881\pm0.040$\tabularnewline
		\midrule 
		AlexNet  & $79.8\%\pm14.0\%$  & $0.672\pm0.401$  & $0.856\pm0.039$\tabularnewline
		\bottomrule
	\end{tabular}
\end{table*}

\subsection{Comparison to Cutting-edge Techniques}

To evaluate the performance of the proposed CNN-SVM-based
model, its results were compared against two leading deep-learning
architectures: GoogLeNet \cite{Szegedy2015} and
AlexNet \cite{Krizhevsky2012}. All three models
were trained and tested on the same roof-defect dataset and used identical
callback configurations to ensure fair comparison. All
	three architectures were configured for the same binary defect/no-defect
	task on identical input sub-images, trained on the same labelled data
	and train/validation/test splits, and evaluated on the same test set.
	The accuracy values reported here reflect architectural differences
	rather than differences in task or data. As illustrated in Fig. \ref{fig:Fig9_AccLossTempor}.(c)
and \ref{fig:Fig9_AccLossTempor}.(d), AlexNet reached
	a training accuracy of approximately $89\%$, and a validation accuracy
	of approximately $83\%$ on the dataset, stopping after $13$ epochs,
	and attained a mean test accuracy of $79.8\%$ over three seeds.
Fig. \ref{fig:Fig9_AccLossTempor}.(e) and \ref{fig:Fig9_AccLossTempor}.(f)
show that GoogLeNet reached a training accuracy of
	approximately $85\%$, and a validation accuracy of approximately
	$87\%$, stopping after $15$ epochs, and attained a mean test accuracy
	of $89.2\%$ over three seeds. In comparison, the proposed CNN-SVM
model (Fig. \ref{fig:Fig9_AccLossTempor}.(a) and \ref{fig:Fig9_AccLossTempor}.(b))
reached a training accuracy of approximately $97.6\%$,
	a validation accuracy of approximately $94.0\%$, and a mean test
	accuracy of $94.4\%$ ($95\%$ CI $\pm0.4\%$ over three seeds),
for more details see Table \ref{tab:modelComparisons}.
A central objective of this research was to develop a model that maintains
high accuracy while minimizing processing time and computational overhead
so that it may be deployed onboard a UAS. Relevant calculations for
operational memory requirements and per-layer computational complexity
are included in the Appendix. Batch-size sensitivity testing was conducted
across all three architectures, with results summarized in Fig. \ref{fig:Fig9_AccLossTempor}.(g).
For the proposed model, processing time scaled approximately
	linearly with batch size, whereas the larger baselines degraded sharply
	at the largest batch. At a batch of $36$ tiles (corresponding with
	a full UAS image) the CNN-SVM model generated predictions in $7.8$
	ms on the RTX~5070 GPU:
\begin{equation}
	\frac{7.8\text{ ms}}{36\text{ tiles}}=\frac{0.22\text{ ms}}{\text{tile}}\label{eq:temp1}
\end{equation}
\begin{figure*}[!t]
	\centering{}\begin{tikzpicture}
		\begin{axis}[width=0.42\textwidth,height=5.4cm,xlabel={Recall},ylabel={Precision},xmin=0,xmax=1,ymin=0.3,ymax=1.02,grid=both,title={(a) Precision--recall}]
			\addplot[blue,thick] table[x=x,y=y] {pr.dat};
			\addplot[red,only marks,mark=*,mark size=2pt] coordinates {(0.920,0.953)};
		\end{axis}
	\end{tikzpicture}\hfill{}\begin{tikzpicture}
		\begin{axis}[width=0.42\textwidth,height=5.4cm,xlabel={False positive rate},ylabel={True positive rate},xmin=0,xmax=1,ymin=0,ymax=1.02,grid=both,title={(b) ROC}]
			\addplot[gray,dashed] coordinates {(0,0) (1,1)};
			\addplot[blue,thick] table[x=x,y=y] {roc.dat};
			\addplot[red,only marks,mark=*,mark size=2pt] coordinates {(0.038,0.920)};
		\end{axis}
	\end{tikzpicture} \caption{\label{fig:prroc}Precision--recall (a; average
		precision $0.980$) and ROC (b; AUC $0.976$) curves on the held-out
		test set, with defect as the positive class. The marked point is the
		default-threshold operating point (recall $0.920$, precision $0.953$,
		false-positive rate $0.038$), consistent with the seed-averaged values
		in Table~\ref{tab:modelComparisons}.}
\end{figure*}
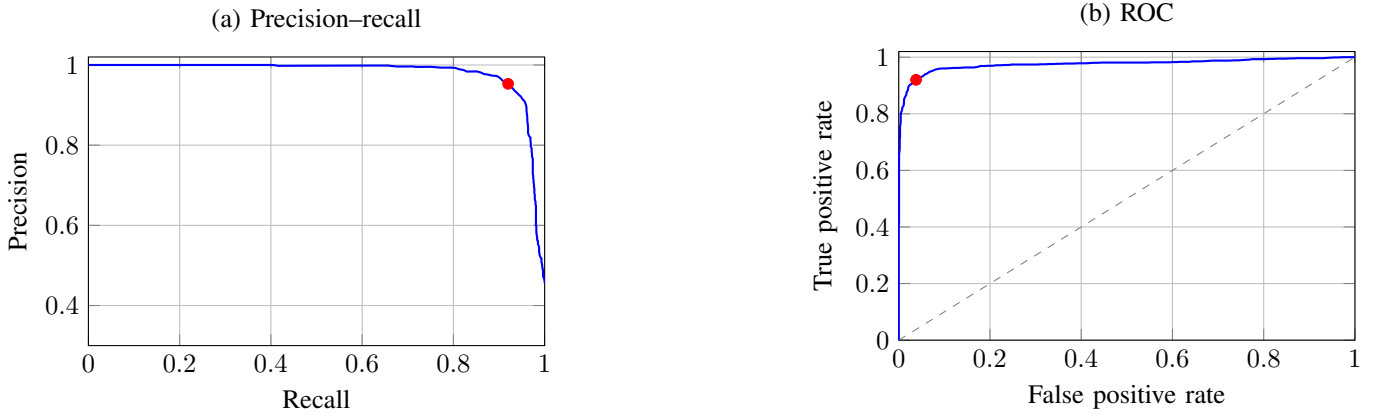
\begin{figure*}[!t]
	\centering{}\includegraphics[width=1\textwidth]{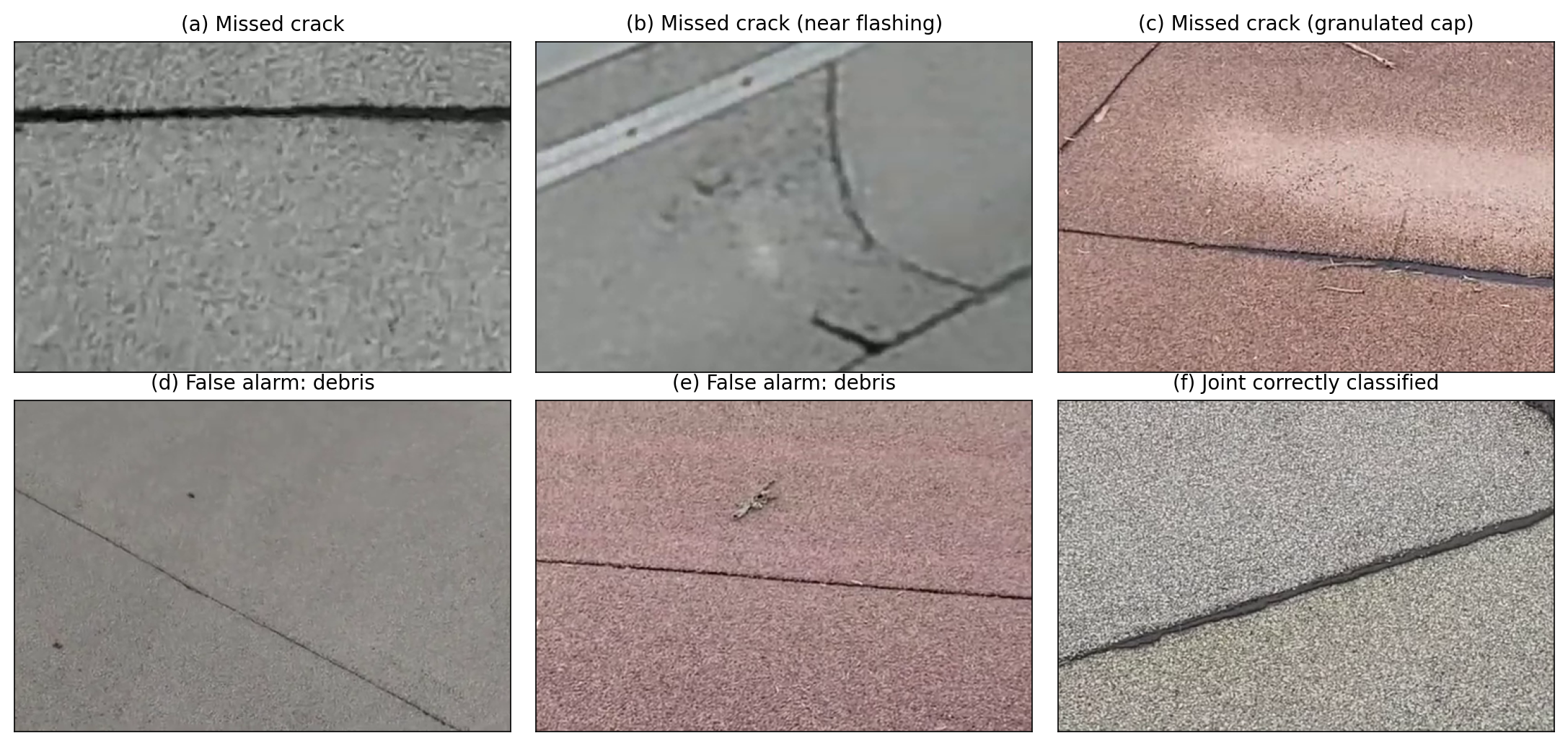}\caption{\label{fig:errorAnalysis}Error analysis on the held-out
		test set. Top row false negatives: thin cracks the model missed, near
		the resolution limit of a tile ((a) on a smooth cap, (b) beside metal
		flashing, (c) on a granulated cap). Bottom row (d) and (e) false positives,
		both triggered by small surface debris (a stone and a twig) rather
		than by membrane joints, and (f) a membrane joint correctly classified
		as non-defective. Installation seams are generally not confused with
		cracks; the residual false positives stem from surface debris.}
\end{figure*}

One complete UAS-captured image is therefore classified
	in well under $10$ ms. The CNN-SVM model was consistently faster
	than GoogLeNet and AlexNet while sustaining the highest accuracy of
	the three architectures: for the same $36$-tile image the processing
	times were $7.8$ ms for CNN-SVM, $22.5$ ms for AlexNet, and $38.9$
	ms for GoogLeNet. The advantage widens as batch sizes increase. At
	a batch of $576$ tiles the CNN-SVM model remained the fastest at
	$117.6$ ms. Conversely, AlexNet slowed to $2,786$ ms under memory
	pressure and GoogLeNet exceeded the available GPU memory. All values
	are averaged over $30$ forward passes per configuration after warm-up.
These results demonstrate that the proposed CNN-SVM
architecture delivers a balance of accuracy and computational efficiency.
This in turn suggests a good suitability for near real-time deployment
in autonomous roof-inspection applications.

\begin{table}[!ht]
	\centering{}\caption{\label{tab:modelParameters}Model parameters, required space, and
		total operations. Memory is shown as an $8$-byte
		upper bound; the deployed FP32 model uses $4$ bytes per parameter
		($\approx446$\,MB total), and the operation counts are FLOPs per
		tile.}
	\begin{tabular}{ll>{\raggedright}p{1.7cm}>{\raggedright}p{1.8cm}}
		\toprule 
		Layers  & Parameters  & Space needed (bytes)  & Total operation performed\tabularnewline
		\midrule
		\midrule 
		Resizing  & 0  & 0  & 0\tabularnewline
		\midrule 
		Random Rotation  & 0  & 0  & 0\tabularnewline
		\midrule 
		Random Brightness  & 0  & 0  & 0\tabularnewline
		\midrule 
		Convolution 2D-1  & 896  & 7,168  & 770,864\tabularnewline
		\midrule 
		Max Pooling 2D-1  & 0  & 0  & 0\tabularnewline
		\midrule 
		Convolution 2D-2  & 18,496  & 147,968  & 1,931,904\tabularnewline
		\midrule 
		Convolution 2D-3  & 73,856  & 590,848  & 3,863,808\tabularnewline
		\midrule 
		Max Pooling 2D-2  & 0  & 0  & 0\tabularnewline
		\midrule 
		Convolution 2D-4  & 295,168  & 2,361,344  & 1,751,040\tabularnewline
		\midrule 
		Convolution 2D-5  & 1,180,160  & 9,441,280  & 3,502,080\tabularnewline
		\midrule 
		Max Pooling 2D-3  & 0  & 0  & 0\tabularnewline
		\midrule 
		Flatten  & 0  & 112,640  & 0\tabularnewline
		\midrule 
		Dense-1  & 115,344,384  & 922,755,072  & 230,687,744\tabularnewline
		\midrule 
		Dense-2  & 131,200  & 1,049,600  & 262,272\tabularnewline
		\midrule 
		Dense-3  & 2,064  & 16,512  & 4,112\tabularnewline
		\midrule 
		Dense-4  & 17  & 136  & 33\tabularnewline
		\midrule 
		Output  & 1  & 8  & 0\tabularnewline
		\bottomrule
	\end{tabular}
\end{table}

\begin{figure*}[!ht]
	\centering{} \definecolor{tpcol}{rgb}{0.89,0.69,0.50} \definecolor{fncol}{rgb}{0.62,0.69,0.81}
	\definecolor{fpcol}{rgb}{0.90,0.87,0.66} \definecolor{tncol}{rgb}{0.65,0.76,0.63}
	\resizebox{\textwidth}{!}{\begin{tikzpicture}[
			cell/.style=rounded corners=3pt, minimum width=2.5cm, minimum height=1.7cm, align=center, font=\bfseries\large, text=black,
			box/.style=rounded corners=5pt, minimum width=2.6cm, minimum height=1.3cm, align=center, font=\bfseries]
			% Panel (a): confusion matrix (seed-42, default-threshold operating point)
			\node[cell, fill=tpcol]                 at (0,0)      {TP\\1065};
			\node[cell, fill=fncol]                 at (2.6,0)    {FN\\92};
			\node[cell, fill=fpcol]                 at (0,-1.8)   {FP\\53};
			\node[cell, fill=tncol]                 at (2.6,-1.8) {TN\\1330};
			\node[font=\bfseries] at (0,1.2)     {Defect};
			\node[font=\bfseries] at (2.6,1.2)   {No Defect};
			\node[font=\bfseries] at (1.3,1.95)  {Predicted};
			\node[font=\bfseries, rotate=90] at (-1.95,0)    {Defect};
			\node[font=\bfseries, rotate=90] at (-1.95,-1.8) {No Defect};
			\node[font=\bfseries, rotate=90] at (-2.7,-0.9)  {Actual};
			\node at (1.3,-2.95) {(a)};
			% Panel (b): performance metrics (seed-42, default threshold; consistent with Table 5)
			\begin{scope}[xshift=8cm]
				\node[box, fill=orange!45] at (0,0)      {FPR\\$0.038$};
				\node[box, fill=gray!35]   at (2.9,0)    {FNR\\$0.080$};
				\node[box, fill=yellow!55] at (5.8,0)    {TNR\\$0.962$};
				\node[box, fill=teal!40]   at (0,-1.8)   {Recall\\$0.920$};
				\node[box, fill=green!45]  at (2.9,-1.8) {Accuracy\\$0.943$};
				\node[box, fill=red!40]    at (5.8,-1.8) {Precision\\$0.953$};
				\node at (2.9,-2.95) {(b)};
			\end{scope}
	\end{tikzpicture}}\caption{\label{fig:confusionMatrix}Model performance metric (a) True Positive
		(TP), False Positive (FP), False Negative (FN), True Negative (TN);
		and (b) False Positive Rate (FPR), False Negative Rate (FNR), True
		Negative Rate (TNR), recall, accuracy and precision of the model.}
\end{figure*}
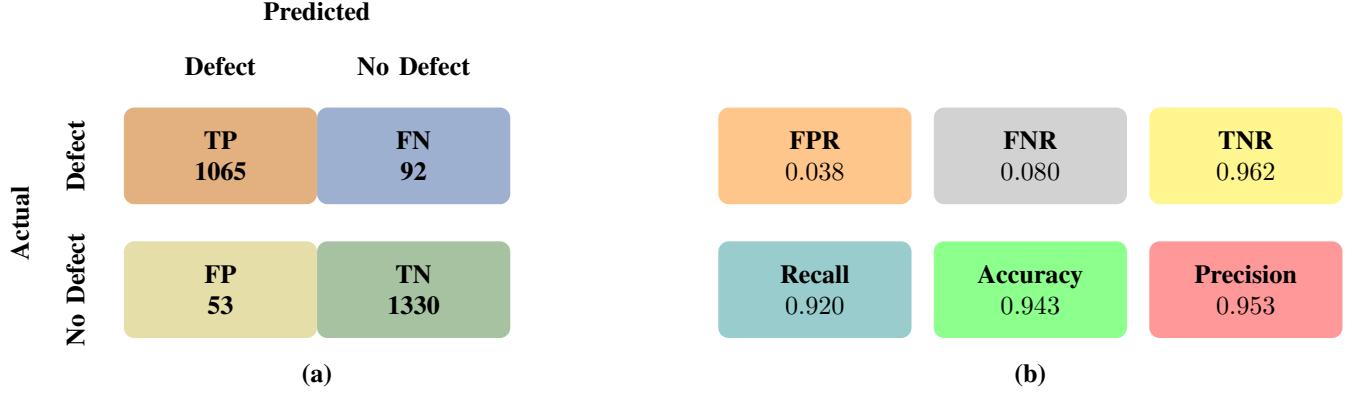

\subsection{CNN-SVM Time Complexities}

The calculations supporting this section are provided in the Appendix.
As illustrated in Fig. \ref{fig:confusionMatrix}, the model's performance
metrics for accuracy, precision, and related indicators are presented.
The confusion matrix in Fig. \ref{fig:confusionMatrix}.(a) reports
the distribution of classifications across the four outcome categories.
Of the $2,540$ test images, $1,065$ were true positives
	(TP), $53$ were false positives (FP), $1,330$ were true negatives
	(TN), and $92$ were false negatives (FN). In total, $2,395$ images
	were correctly classified and $145$ were misclassified. The model,
while not perfect, particularly with respect to the
	small fraction of missed defects (false negatives), the overall classification
accuracy remains high and is considered successful for the objectives
of this research.

\vspace{0.2cm}
A missed defect (false negative) is considered the
	most significant since it can lead to water ingress, mould growth,
	or structural deterioration. Therefore, recall and the false-negative
	rate are emphasized instead of overall accuracy for this safety-critical
	task. On a class-wise basis, of the $1,157$ defective tiles in the
	test set the model correctly identified $1,065$ (a recall of $92.0\%$)
	and missed $92$ (a false-negative rate of $8.0\%$). Of the non-defective
	times ($1,383$) the model correctly identified $1,330$ (a true-negative
	rate of $96.2\%$) and produced only $53$ false alarms (a false-positive
	rate of $3.8\%$). Therefore, the model combines a high recall with
	a low false-alarm rate. Since a missed defect is the costliest outcome,
	the decision threshold can be lowered to trade precision for additional
	recall whenever an application demands it. From the precision-recall
	and ROC curves for the test set (Fig.~\ref{fig:prroc}) the model
	reports an average precision of $0.980$ and an ROC-AUC of $0.976$.
	These values are consistent with the recall and precision values reported
	in Table~\ref{tab:modelComparisons}.

\vspace{0.2cm}
On the test set, the CNN-SVM model achieved a mean
	test accuracy of $94.4\%$ ($95\%$ confidence interval $\pm0.4\%$
	over three seeds). It therefore clearly outperformed GoogLeNet ($89.2\%$)
	and AlexNet ($79.8\%$). Additional performance indicators are presented
in Fig. \ref{fig:confusionMatrix}.(b), which demonstrate that the
CNN-SVM model maintains high precision, recall,
and overall predictive reliability. Consequently, the CNN-SVM
architecture used in this study performs comparably to other state-of-the-art
machine learning models.

\vspace{0.2cm}
To characterize the model's errors the misclassified
	test tiles were collected and analyzed (Fig.~\ref{fig:errorAnalysis}).
	The false negatives are dominated by thin or hairline cracks near
	the resolution limit of a tile (Fig.~\ref{fig:errorAnalysis}(a)--(c)):
	the model reliably detects prominent cracking and alligatoring but
	can miss faint fractures, consistent with the $8.0\%$ false-negative
	rate. The false positives are driven by loose surface debris such
	as a small stone or a twig on otherwise-sound membrane (Fig.~\ref{fig:errorAnalysis}
	(d)--(e)). It is believed that with additional training data, debris
	within an image will become less impactful and easier for the model
	to ignore. Membrane seams, which are present throughout the dataset,
	are in general classified correctly as non-defective (Fig.~\ref{fig:errorAnalysis}(f)).
	Overall, the model appears to focus on the irregular features of a
	dark fracture pattern of a crack rather than the regular straight
	line of a seam. Throughout the false classifications, no systematic
	tendency to mistake assembly gaps between membranes as defects was
	observed.

\vspace{0.2cm}
Table \ref{tab:modelParameters} also lists the number of operations
executed within each layer, with a cumulative total of $242,773,857$
operations. This metric is useful for evaluating processor capability
when selecting an appropriate UAS platform to ensure the system can
maintain the required per-second throughput. The operational demand
remains well below the typical processing capacity of enterprise-grade
UAS systems \cite{hash2025_RIENG_Avionics}, and thus does not constitute
a significant deployment limitation. All calculations supporting these
values are based on the equations provided in Appendix~\ref{app:calcs}.

\vspace{0.2cm}
The overall compute profile of the model is summarized
	here. The network produced by this research has approximately $117$
	million parameters. This is primarily dominated by the first dense
	layer which has ($115$ million). With single precision (FP32, $4$
	bytes per parameter) this is approximately $446$\,MB of weights.
	Each tile therefore requires about $243$ million floating-point operations
	for classification. A full $36$-tile image requires about $8.7$
	billion operations. The peak activation memory is on the order of
	a few megabytes per tile and is negligible relative to the weights.
	Inference used FP32 with a batch size of $32$ tiles, giving the per-tile
	and per-image latencies reported above. These inference costs are
	different from the one-time training cost ($17$\,h $15$\,m on
	the RTX~5070). On-hardware profiling on an embedded accelerator is
	identified as future work in the Limitations and Future Work subsection.

\subsection{Limitations and Future Work}

Several limitations define the scope of this study
	and motivate future work. Firstly, the training and test sets were
	divided at the level of whole source photographs. Since consecutive
	UAS images typically have a substantial overlap, many photos share
	pixels of the same roof which means that the same data can appear
	across training, validation and test sets. If the data was separated
	in this fashion, a slight decrease in test accuracy would be expected.
	A roof- or flight-level split that keeps all images of a given roof
	within a single set is planned for future development in order to
	quantify and remove this effect. \textcolor{black}{Splitting the dataset
		at the photo level elimiates tile and augmentation level leakage since
		all tiles that are derived from a single photograph are confined to one
		set. However, leakage can still arise on the roof or flight level since
		different photographs may contain spatial overlaps of the same region with
		other photgraphs. Consecutive photographs of the same roof will often
		contain portions of the same spatial area and the current data split
		does not eliminate this. Consequently, the reported accuracy must be
		considered with only modest optimism until a complete leakage-free
		split is tested against.} Secondly, although the model was
	designed for embedded operation, all training and inference metrics
	reported here were performed on a workstation GPU. The proposed model
	executes approximately $243$ million operations per tile (about $8.7$
	billion per full $36$-tile image) and requires under $1$\,GB of
	parameter memory. These characteristics are within the envelope of
	current embedded accelerators such as the NVIDIA Jetson Orin Nano
	(up to $67$ TOPS at $7$--$25$\,W) and AGX Orin (up to $275$
	TOPS)~\cite{NvidiaJetsonOrin2022}. However, onboard deployment,
	as well as the profiling of power, latency and thermal behaviour remain
	future work. Thirdly, the reported metrics come from a small number
	of training runs, using seeds $42$, $43$, and $44$. For future
	testing, additional repeated runs are planned to further characterize
	variance. Fourth, images were excluded from this dataset if they contained
	walls, vehicles, or mechanical units. This was done in order to provide
	a clean set of training data. For real-world application of this technology,
	it will be important to incorporate these images into future models
	in order to maintain applicability and effective range. Additionally,
	the benefit of the dual-altitude data collection methodology is motivated
	by prior multi-altitude inspection studies~\cite{Fan2025}. However,
	it has not yet been isolated through a dedicated ablation. Finally,
	performance under specific environmental conditions such as wet, dry,
	shaded, or bright areas or across different roofing-membrane systems
	was not evaluated. This alone warrants a deep study and additional
	research for integration with future models.

\begin{table*}[!ht]
	\caption{\label{tab:metricPerformance}Metrics performance and the related
		mathematical formulas}
	
	\centering{}%
	\begin{tabular}{ccp{9cm}}
		\toprule 
		Metrics Performance  & Mathematical formula  & Interpretation\tabularnewline
		\midrule
		\midrule 
		False Positive Rate (FPR)  & $\frac{FP}{FP+TN}$  & Proportion of non-defective images incorrectly classified
		as defective\tabularnewline
		\midrule 
		False Negative Rate (FNR)  & $\frac{FN}{TP+FN}$  & Proportion of defective images incorrectly classified
		as non-defective\tabularnewline
		\midrule 
		True Negative Rate (TNR)  & $\frac{TN}{FP+TN}$  & Proportion of non-defective images correctly classified
		as non-defective\tabularnewline
		\midrule 
		Recall  & $\frac{TP}{TP+FN}$  & Proportion of defective images correctly classified
		as defective (sensitivity)\tabularnewline
		\midrule 
		Accuracy  & $\frac{TP+TN}{TP+FN+TN+FP}$  & Measure of the model's correct classification rate\tabularnewline
		\midrule 
		Precision  & $\frac{TP}{TP+FP}$  & Proportion of images predicted defective that are
		truly defective\tabularnewline
		\bottomrule
	\end{tabular}
\end{table*}

\section{Conclusion\label{sec:conclusion}}

This research developed a lightweight, high-performance model
capable of supporting autonomous inspections of flat roofing membrane
systems. Designed specifically for deployment on Unmanned Aerial Systems
(UASs), the model achieves high predictive accuracy while remaining
computationally efficient, an essential requirement given the limited
onboard processing capabilities of contemporary UAS platforms. The
	tiling-based neural architecture decomposes high-resolution roof images
	into smaller, computationally manageable tiles. These tiles
enable rapid inference without compromising the level of detail required
for reliable defect detection. A significant contribution of this
work is the creation of a bespoke dataset, addressing the absence
of publicly available datasets for flat-roof defect detection. Experimental
validation was achieved through images collected
mainly through onsite UAS-based field surveys (DJI Matrice 350 RTK
drone) and supplemented with online sources, then augmented to expand
the dataset size and enhance environmental and visual diversity. Several
modelling strategies including data augmentation layers, callback
mechanisms to mitigate overfitting, and the use of a
Stochastic Gradient Descent (SGD) optimizer with momentum, were employed
to achieve strong performance metrics. The final model experimental
assessment attained a training accuracy of approximately
	$97.6\%$, a validation accuracy of approximately $94.0\%$, and a
	mean test accuracy of $94.4\%$ ($95\%$ CI $\pm0.4\%$ over three
	seeds). When benchmarked against established architectures such as
AlexNet and GoogLeNet, the proposed CNN-SVM model
demonstrated superior accuracy while requiring substantially
less processing time. The tiling strategy, which
reduces full-resolution images ($3,840\times2,160$ pixels) into $640\times360$-pixel
sectors, enabled the model to process a full $36$-tile
	image in roughly $7.8$\,ms, about $130$ full images per second.
	This far exceeds the target throughput of one full image per second
and suggests that further architectural refinements may yield even
greater processing efficiency. Beyond model development, this research
also outlined a complete operational workflow for UAS-based roof inspections,
including image acquisition, tiling, classification,
and operator feedback. The workflow maintains an effective balance
between computational demand and real-time performance while preserving
capacity for critical UAS functions such as navigation, communication,
and collision avoidance.

\section*{CRediT authorship contribution statement}

\textbf{Samuel Dunthorne}: Writing - original draft, Visualization,
Validation, Software, Methodology, Investigation, Formal analysis,
Conceptualization.

\textbf{Hashim A. Hashim}: Writing - review and editing, Visualization,
Validation, Supervision, Methodology, Investigation, Funding acquisition,
Conceptualization.

\section*{Declaration of competing interest}

The authors declare that they have no known competing financial interests
or personal relationships that could have appeared to influence the
work reported in this paper.

\section*{Data availability}

The dataset 'tiled and augmented UAS imagery of modified-bitumen flat-roof defects'
	together with the reproducible train/validation/test split index (\texttt{manifest.csv}), is openly 
	available on Zenodo at \url{https://doi.org/10.5281/zenodo.20594605} (CC~BY~4.0).The training, 
	evaluation, tiling, augmentation, and dataset splitting scripts are openly archived on Zenodo at 
	\url{https://doi.org/10.5281/zenodo.20599363}. The trained model weights are archived at 
	\url{https://doi.org/10.5281/zenodo.20596485}. All experiments use a fixed random seed ($42$).

\appendices{}

\section{Calculation of Performance and Time Complexities\label{app:calcs}}

When evaluating the model's performance, True Positives (TP) represent
the number of images correctly classified as defective, while True
Negatives (TN) denote images correctly identified as non-defective.
False Positives (FP) correspond to non-defective images incorrectly
classified as defective, and False Negatives (FN) refer to defective
images incorrectly classified as non-defective. These metrics, along
with their corresponding calculations, are presented in Table \ref{tab:metricPerformance}
and illustrated in Fig. \ref{fig:confusionMatrix}. To determine the
number of trainable parameters in each layer, the following formulations
were applied. For a convolutional layer, where $w\times h$ is the
filter size, $c$ is the number of input channels, and $k$ is the
number of filters, the total number of parameters is defined in Eq.
\eqref{eq:Param1}:
\begin{equation}
	\text{\# of Parameters}=(w\times h\times c+1)\times k\label{eq:Param1}
\end{equation}
For a dense (fully connected) layer, where $n$ denotes the number
of input units and $m$ denotes the number of output units, the number
of parameters is computed in Eq. \eqref{eq:Param2} as
\begin{equation}
	\text{\# of Parameters}=(n\times m)+m\label{eq:Param2}
\end{equation}
The total number of operations required per convolutional layer was
calculated using Eq. \eqref{eq:TotalOps} as below
\begin{equation}
	\text{Total Ops}=\frac{2(c\times w\times h)(M-w+s)(N-h+s)}{s^{2}}\label{eq:TotalOps}
\end{equation}
where $M$ and $N$ represent the spatial dimensions of the input
feature map and $s$ is the stride length. Similarly, the total number
of operations for a dense layer was determined using
\begin{equation}
	\text{Total Ops}=m(2n+1)\label{eq:TotalOps2}
\end{equation}
where $n$ and $m$ in Eq. \eqref{eq:TotalOps} retain their definitions
as the number of input and output units, respectively.

\section*{Acknowledgement}

This work was supported in part by Mitacs through the Mitacs Accelerate
Program and Roofmaster Ottawa Inc, under grant number IT42196.

\ifCLASSOPTIONcaptionsoff \fi

\bibliographystyle{ieeetr}
\bibliography{bib_RoofDefect}

@article{seo2025quantitative,
  title={Optimizing urban infrastructure resilience: Analyzing cascading failures and critical node dependencies through multilayer network models},
author={Lu, Cong and et al.},
journal={J. of Safety Science and Resilience},
pages={100245},
year={2025},
publisher={Elsevier}
}

@article{yiugit2024automatic,
  title={Maritime man-overboard search using a lightweight and efficient end-to-end detection transformer},
author={Xu, Guokang and Yin, Jianchuan and Wang, Nini and Zhang, Zeguo},
journal={Journal of Safety Science and Resilience},
pages={100267},
year={2025},
publisher={Elsevier}
}

@article{wang2025study,
  title={Influencing Factors and Mechanisms of Super High-Rise Buildings Safety Risks: A Fuzzy-DEMATEL-AISM Analysis},
author={Jia, Ziyu and et al.},
journal={Journal of Safety Science and Resilience},
pages={100232},
year={2025},
publisher={Elsevier}
}

@techReport{NaturalResourcesCanada2025,
	title = {Cooling and {V}entilating {E}quipment for {R}esidential {U}se},
	month = {3},
	institution = {Natural Resources Canada, Government of Canada},
	url   = {https://natural-resources.canada.ca/energy-efficiency/energy-star/products/list-certified-products/cooling-ventilating-equipment-residential-use},
	year = {2025}
}

@techReport{HealthCanada2023,
	title = {{G}uide to {A}ddressing {M}oisture and {M}ould {I}ndoors},
	institution = {Health Canada, Government of Canada},
	month = {1},
	year = {2023},
	url = {https://www.canada.ca/en/health-canada/services/publications/healthy-living/addressing-moisture-mould-your-home.html}
}

@techReport{EnvironmentandClimateChangeCanada2025,
	title = {{C}anadian {E}nvironmental {S}ustainability {I}ndicators: {G}reenhouse {G}as {E}missions},
	institution = {Environment and Climate Change Canada, Government of Canada},
	year = {2025},
	month = {3},
	isbn = {9780660753379},
	pages = {28},
	url = {https://www.canada.ca/en/environment-climate-change/services/environmental-indicators/greenhouse-gas-emissions.html}
}

@techReport{CanadaEnergyRegulator2023,
	title = {{E}lectricity {G}eneration},
	institution = {Canada Energy Regulator, Government of Canada},
	year = {2023},
	howpublished = {Canada's Energy Future Data Appendices},
	url = {https://doi.org/10.35002/zjr8-8x75}
}

@techReport{GovernmentOfOntario2025,
	title = {{T}he {E}nd of {C}oal},
	institution = {Ministry of Energy and Mines Government of Ontario, Government of Canada},
	month = {10},
	year = {2025},
	url = {https://www.ontario.ca/page/end-coal}
}

@techReport{MinistryofHousing2024,
	title = {{E}nglish {H}ousing {S}urvey 2022 to 2023: {H}ousing {Q}uality and {C}ondition},
	institution = {Ministry of Housing, Communities and Local Government, Government of the United Kingdom},
	month = {7},
	year = {2024},
	url = {https://www.gov.uk/government/statistics/english-housing-survey-2022-to-2023-housing-quality-and-condition}
}

@techReport{InsuranceInformationInstitute2025,
	title = {{S}potlight on {C}atastrophes - {I}nsurance {I}ssues},
	institution = {Insurance Information Institute},
	month = {2},
	year = {2025},
	url = {https://www.iii.org/article/spotlight-on-catastrophes-insurance-issues}
}

@techReport{StatisticsCanada2025,
	title = {{C}anadian {H}ousing {S}tatistics {P}rogram (CHSP)},
	institution = {Statistics Canada, Government of Canada},
	month = {10},
	year = {2025},
	url = {https://www.statcan.gc.ca/imdb-bmdi/5257-eng.htm}
}

@techReport{StatisticsFinland2025,
	title = {{D}wellings and {H}ousing {C}onditions},
	institution = {Statistics Finland, Government of Finland},
	month = {10},
	year = {2025},
	url = {https://stat.fi/en/statistics/asas}
}

@techReport{StatisticsIceland2023,
	title = {{P}roportion of {R}ental {D}wellings {D}ecreased between 2011 and 2021},
	institution = {Statistics Iceland, Government of Iceland},
	month = {12},
	year = {2023},
	url = {https://www.statice.is/publications/news-archive/census/census-2021-housing/}
}

@techReport{StatisticsNorway2025,
	title = {{B}uilding {S}tock},
	institution = {Statistics Norway, Government of Norway},
	month = {2},
	year = {2025},
	url = {https://www.ssb.no/en/bygg-bolig-og-eiendom/bygg-og-anlegg/statistikk/bygningsmassen}
}

@techReport{StatisticsDenmark2025,
	title = {{H}ouseholds and {F}amilies},
	institution = {Statistics Denmark, Government of Denmark},
	month = {1},
	year = {2025},
	url = {https://www.dst.dk/en/Statistik/emner/borgere/husstande-og-familieforhold/husstande-og-familier}
}

@techReport{StatisticsSweden2025,
	title = {{D}welling {S}tock},
	institution = {Statistics Sweden, Government of Sweden},
	month = {4},
	year = {2025},
	url = {https://www.scb.se/en/finding-statistics/statistics-by-subject-area/housing-construction-and-building/housing-and-accommodation/dwelling-stock/}
}

@techReport{USCensusBureau2024,
	author = {U.S. Census Bureau},
	title = {Table 4. {E}stimates of the {H}ousing {I}nventory},
	institution = {U.S. Census Bureau},
	month = {10},
	year = {2024},
	url = {https://www.census.gov/housing/hvs/files/qtr324/tab4.xlsx}
}

@article{Tariku2023,
  title={Thermal performance of flat roof insulation materials: A review of temperature, moisture and aging effects},
author={Tariku, Fitsum and Shang, Yina and Molleti, Sudhakar},
journal={Journal of Building Engineering},
volume={76},
pages={107142},
year={2023},
publisher={Elsevier}
}

@article{Grant2017,
  title={The influence of roof reflectivity on adjacent air and surface temperatures},
author={Grant, Elizabeth J and Black, Kenneth A and Werre, Stephen R},
journal={Architectural Science Review},
volume={60},
number={2},
pages={137--144},
year={2017},
publisher={Taylor \& Francis}
}

@incollection{Brook2014,
title={The state of air quality in Canada: national patterns},
author={Brook, Jeffrey R and et al.},
booktitle={Air quality management: Canadian perspectives on a global issue},
pages={43--67},
year={2013},
publisher={Springer}
}

@article{Wang2016,
  title={Comparing the effects of urban heat island mitigation strategies for Toronto, Canada},
author={Wang, Yupeng and Berardi, Umberto and Akbari, Hashem},
journal={Energy and buildings},
volume={114},
pages={2--19},
year={2016},
publisher={Elsevier}
}

@article{Gaur2018,
  title={Analysis and modelling of surface Urban Heat Island in 20 Canadian cities under climate and land-cover change},
author={Gaur, Abhishek and Eichenbaum, Markus Kalev and Simonovic, Slobodan P},
journal={Journal of environmental management},
volume={206},
pages={145--157},
year={2018},
publisher={Elsevier}
}

@article{MansouriKouhestani2019,
  title={Evaluating solar energy technical and economic potential on rooftops in an urban setting: the city of Lethbridge, Canada},
author={Mansouri Kouhestani, Fariborz and et al.},
journal={International Journal of Energy and Environmental Engineering},
volume={10},
number={1},
pages={13--32},
year={2019},
publisher={Springer}
}

@article{Pisello2013,
  title={Active cool roof effect: impact of cool roofs on cooling system efficiency},
author={Pisello, Anna Laura and Santamouris, Mattheos and Cotana, Franco},
journal={Advances in building energy research},
volume={7},
number={2},
pages={209--221},
year={2013},
publisher={Taylor \& Francis}
}

@article{Bartels2024,
title={Life Cycle Assessment and Maintenance Planning of an Innovative Flat Roof Solution},
author={Bartels, Giovanna and Flores-Colen, Ines and Silvestre, Jose Dinis and Silva, Luis},
journal={Procedia Structural Integrity},
volume={55},
pages={88--95},
year={2024},
publisher={Elsevier}
}

@article{Li2014,
  title={The effectiveness of cool and green roofs as urban heat island mitigation strategies},
author={Li, Dan and Bou-Zeid, Elie and Oppenheimer, Michael},
journal={Environmental Research Letters},
volume={9},
number={5},
pages={055002},
year={2014},
publisher={IOP Publishing}
}

@article{Dong2013,
title={Fatal falls from roofs among US construction workers},
author={Dong, Xiuwen Sue and Choi, Sang D and Borchardt, James G and Wang, Xuanwen and Largay, Julie A},
journal={Journal of Safety Research},
volume={44},
pages={17--24},
year={2013},
publisher={Elsevier}
}

@article{Aggarwal2024,
  title={A comprehensive review of life cycle assessment (LCA) studies in roofing industry: current trends and future directions},
author={Aggarwal, Chetan and Molleti, Sudhakar and Ghobadi, Mehdi},
journal={Smart Cities},
volume={7},
number={5},
pages={2781--2801},
year={2024},
publisher={MDPI}
}

@article{Zahradnik2024,
  title={Deep Roof Refiner: A detail-oriented deep learning network for refined delineation of roof structure lines using satellite imagery},
author={Qian, Zhen and et al.},
journal={International Journal of Applied Earth Observation and Geoinformation},
volume={107},
pages={102680},
year={2022},
publisher={Elsevier}
}

@article{Santos2023,
  title={Deep learning applied to equipment detection on flat roofs in images captured by UAV},
author={dos Santos, Lara Monalisa Alves and et al.},
journal={Case Studies in Construction Materials},
volume={18},
pages={e01917},
year={2023},
publisher={Elsevier}
}

@article{Santos2025,
  title={Mapping stains on flat roofs using semantic segmentation based on deep learning},
author={dos Santos, Lara Monalisa Alves and et al.},
journal={Case Studies in Construction Materials},
volume={22},
pages={e04106},
year={2025},
publisher={Elsevier}
}

@article{Fan2025,
  title={{UAV} and Deep Learning for Automated Detection and Visualization of Fa\c{c}ade Defects in Existing Residential Buildings},
author={Fan, Yue and et al.},
journal={Sensors},
volume={25},
number={23},
pages={7118},
year={2025},
publisher={MDPI}
}

@article{Shorten2019,
  title={A survey on Image Data Augmentation for Deep Learning},
author={Shorten, Connor and Khoshgoftaar, Taghi M},
journal={Journal of Big Data},
volume={6},
pages={60},
year={2019},
publisher={Springer}
}

@misc{NvidiaJetsonOrin2022,
title={{J}etson {AGX} {O}rin {S}eries {T}echnical {B}rief},
author={{NVIDIA Corporation}},
howpublished={NVIDIA Technical Brief v1.2},
year={2022}
}

@inproceedings{Jiao2019,
 title={A deep learning based forest fire detection approach using {UAV} and YOLOv3},
author={Jiao, Zhentian and et al.},
booktitle={2019 1st International conference on industrial artificial intelligence (IAI)},
pages={1--5},
year={2019},
organization={IEEE}
}

@phdthesis{Sakerka2012,
  title={Evaluating Strategies for Wide Scale Replacement of Human Inspection with Machine Vision},
author={Sakerka, Lauren},
year={2022},
school={Massachusetts Institute of Technology}
}

@article{Ren2015,
  title={Faster R-CNN: Towards real-time object detection with region proposal networks},
author={Ren, Shaoqing and He, Kaiming and Girshick, Ross},
journal={IEEE transactions on pattern analysis and machine intelligence},
volume={39},
number={6},
pages={1137--1149},
year={2016},
publisher={IEEE}
}

@article{Bochkovskiy2020,
  title={Yolov4: Optimal speed and accuracy of object detection},
author={Bochkovskiy, Alexey and Wang, Chien-Yao and Liao, Hong-Yuan Mark},
journal={arXiv preprint arXiv:2004.10934},
year={2020}
}

@book{Guan2019,
  title={Image processing and acquisition using Python},
author={Chityala, Ravishankar and Pudipeddi, Sridevi},
year={2020},
publisher={Chapman and Hall/CRC}
}

@incollection{Wang2008,
  title={Optimized scale-and-stretch for image resizing},
author={Wang, Yu-Shuen and Tai, Chiew-Lan and Sorkine, Olga and Lee, Tong-Yee},
booktitle={ACM SIGGRAPH Asia 2008 papers},
pages={1--8},
	year = {2008},
	month = {12},
	publisher = {ACM},
	city = {New York, NY, USA}
}

@article{Kim2017,
title={Recent advances in convolutional neural networks},
author={Gu, Jiuxiang and et al.},
journal={Pattern recognition},
volume={77},
pages={354--377},
year={2018},
publisher={Elsevier}
}

@article{Conceicao2017,
  title={Inspection, diagnosis, and rehabilitation system for flat roofs},
author={Conceicao, J and et al.},
journal={Journal of Performance of Constructed Facilities},
volume={31},
number={6},
pages={04017100},
year={2017},
publisher={American Society of Civil Engineers}
}

@article{Carretero-Ayuso2016,
  title={Analysis of the execution deficiencies of flat roofs with bituminous membranes},
author={Carretero-Ayuso, Manuel J and De Brito, Jorge},
journal={Journal of Performance of Constructed Facilities},
volume={30},
number={6},
pages={04016049},
year={2016},
publisher={American Society of Civil Engineers}
}

@inproceedings{Agarap2019,
  title={Image Classification using SVM and CNN},
author={Chaganti, Sai Yeshwanth and et al.},
booktitle={2020 International conference on computer science, engineering and applications (ICCSEA)},
pages={1--5},
year={2020},
organization={IEEE}
}

@article{Cortes1995,
  title={Support-vector networks},
author={Cortes, Corinna and Vapnik, Vladimir},
journal={Machine learning},
volume={20},
number={3},
pages={273--297},
year={1995},
publisher={Springer}
}

@article{Krizhevsky2012,
  title={Imagenet classification with deep convolutional neural networks},
author={Krizhevsky, Alex and Sutskever, Ilya and Hinton, Geoffrey E},
journal={Advances in neural information processing systems},
volume={25},
year={2012}
}

@inproceedings{Szegedy2015,
  title={Going deeper with convolutions},
author={Szegedy, Christian and et al.},
booktitle={Proceedings of the IEEE Conference on Computer Vision and Pattern Recognition (CVPR)},
pages={1--9},
year={2015}
}

@article{Tang2015,
  title={Deep learning using linear support vector machines},
author={Tang, Yichuan},
journal={arXiv preprint arXiv:1306.0239},
year={2013}
}

@inproceedings{Poojary2019,
  title={Comparative study of model optimization techniques in fine-tuned CNN models},
author={Poojary, Ramaprasad and Pai, Akul},
booktitle={2019 International Conference on Electrical and Computing Technologies and Applications (ICECTA)},
pages={1--4},
year={2019},
organization={IEEE}
}

@article{Bouguettaya2022,
  title={A review on early wildfire detection from unmanned aerial vehicles using deep learning-based computer vision algorithms},
author={Bouguettaya, Abdelmalek and et al.},
journal={Signal Processing},
volume={190},
pages={108309},
year={2022},
publisher={Elsevier}
}

@article{Bhatt2021,
 title={Image-based surface defect detection using deep learning: A review},
author={Bhatt, Prahar M and et al.},
journal={Journal of Computing and Information Science in Engineering},
volume={21},
number={4},
pages={040801},
year={2021},
publisher={American Society of Mechanical Engineers}
}

@article{Tulbure2022,
  title={A review on modern defect detection models using DCNNs--Deep convolutional neural networks},
author={Tulbure, Andrei-Alexandru and Tulbure, Adrian-Alexandru and Dulf, Eva-Henrietta},
journal={Journal of Advanced Research},
volume={35},
pages={33--48},
year={2022},
publisher={Elsevier}
}

@article{Ling2023,
  title={An autoencoder with convolutional neural network for surface defect detection on cast components},
author={Chamberland, Olivia and Reckzin, Mark and Hashim, Hashim A},
journal={Journal of Failure Analysis and Prevention},
volume={23},
number={4},
pages={1633--1644},
year={2023},
publisher={Springer}
}

@article{Gullbrekken2016,
  title={Process induced building defects in Norway--development and climate risks},
author={Bunkholt, Nora Schjoth and Gullbrekken, Lars and Time, Berit and Kvande, Tore},
journal={Journal of Physics: Conference Series},
volume={2069},
pages={012040},
year={2021},
organization={IOP publishing}
}

@article{hash2026_MFWAV,
	title={{F}rom {I}nsects to {B}io-inspired {M}icro {F}lapping {W}ing {A}erial {V}ehicles {I}ntelligent {F}light: {A} {R}eview, {D}esign {P}rinciples, and {F}uture {P}rospects},
	author={Hashim, Hashim A},
	journal={Digital Engineering},
	pages={100114},
	year={2026},
	publisher={Elsevier}
}

@article{hash2025_RIENG_Avionics,
	title={{A}dvances in {UAV} {A}vionics {S}ystems {A}rchitecture, {C}lassification and {I}ntegration: {A} {C}omprehensive {R}eview and {F}uture {P}erspectives},
	author={Hashim, Hashim A},
	journal={Results in Engineering},
	pages={103786},
	volume={25},
	year={2025},
	publisher={Elsevier}
}

@article{Aditya2024_SegNet,
	title={{S}eg{N}et: {A} {S}egmented {D}eep {L}earning based {C}onvolutional {N}eural {N}etwork {A}pproach for {D}rones {W}ildfire {D}etection},
	author={Jonnalagadda, Aditya V and Hashim, Hashim A},
	journal={Remote Sensing Applications: Society and Environment},
	volume={34},
	pages={101181},
	year={2024},
	publisher={Elsevier}
}

@INPROCEEDINGS{Aditya_2024_ICDS_comprehensive,
	title={{C}omprehensive and {C}omparative {A}nalysis between {T}ransfer {L}earning and {C}ustom {B}uilt {VGG} and {CNN}-{SVM} {M}odels for {W}ildfire {D}etection},
	author={Jonnalagadda, Aditya V and Hashim, Hashim A and Harris, Andrew},
	booktitle={2024 IEEE Sixth International Conference on Intelligent Computing in Data Sciences (ICDS)},
	pages={1--7},
	year={2024},
	organization={IEEE}
}

@article{hash2023_ISA_UAV_NavCont,
	title={{O}bserver-based {C}ontroller for {VTOL}-{UAV}s {T}racking using {D}irect {V}ision-Aided {I}nertial {N}avigation {M}easurements},
	author={Hashim, Hashim A and Eltoukhy, Abdelrahman EE and Odry, Akos},
	journal={ISA transactions},
	volume={137},
	pages={133--143},
	year={2023},
	publisher={Elsevier}
}

@article{hash2023_IJC_UAV_NavCont,
	title={{E}xponentially {S}table {O}bserver-based {C}ontroller for {VTOL}-{UAV}s without {V}elocity {M}easurements},
	author={Hashim, Hashim A},
	journal={International Journal of Control},
	volume={96},
	number={8},
	pages={1946--1960},
	year={2023},
	publisher={Taylor \& Francis}
}

\end{document}